\documentclass[11pt,a4paper]{article}
\usepackage[hyperref]{emnlp2020} 
\usepackage{makecell}
\usepackage{hyperref}
\usepackage{graphicx}
\usepackage{caption}
\usepackage{subcaption}
\usepackage{latexsym}

\usepackage{bbm}
\usepackage{amsmath,amsfonts,amsthm,amssymb} 
\usepackage{bm}
\usepackage{multirow}
\usepackage{microtype}
\usepackage{graphicx}

\aclfinalcopy 

\title{Hierarchical Pre-training for Sequence Labelling in Spoken Dialog}

\author{Emile Chapuis\textsuperscript{\rm 1}\footnote[1]{Equal contribution}, \textbf{Pierre Colombo\textsuperscript{\rm 1,2}\thanks{\rm stands for equal contribution}, Matteo Manica\textsuperscript{\rm 3}}\\ \textbf{Matthieu Labeau\textsuperscript{\rm 1}, Chloe Clavel\textsuperscript{\rm 1}}\\
\textsuperscript{\rm 1}LTCI, Telecom Paris, Institut Polytechnique de Paris, \textsuperscript{\rm 2}IBM GBS France, \textsuperscript{\rm 3}IBM Research Zurich\\
\textsuperscript{\rm 1}firstname.lastname@telecom-paris.fr, \textsuperscript{\rm 3}tte@zurich.ibm.com 
}

\date{}

\begin{document}
\maketitle
\begin{abstract}
Sequence labelling tasks like Dialog Act and Emotion/Sentiment identification are a key component of spoken dialog systems. In this work, we propose a new approach to learn generic representations adapted to spoken dialog, which we evaluate on a new benchmark we call Sequence labellIng evaLuatIon benChmark fOr spoken laNguagE benchmark (\texttt{SILICONE}). \texttt{SILICONE}\footnote{Benchmark can be found in the dataset library from HuggingFace \cite{2020HuggingFace-datasets} at \url{https://huggingface.co/datasets/silicone}} is model-agnostic and contains 10 different datasets of various sizes. We obtain our representations with a hierarchical encoder based on transformer architectures, for which we extend two well-known pre-training objectives. Pre-training is performed on OpenSubtitles: a large corpus of spoken dialog containing over $2.3$ billion of tokens.
We demonstrate how hierarchical encoders achieve competitive results with consistently fewer parameters compared to state-of-the-art models and we show their importance for both pre-training and fine-tuning.
\end{abstract}

\section{Introduction}




The identification of both Dialog Acts (\texttt{DA}) and Emotion/Sentiment (\texttt{E/S}) in spoken language is an important step toward improving model performances on spontaneous dialogue task.
Especially, it is essential to avoid the generic response problem, i.e., having an automatic dialog system generate an unspecific response --- that can be an answer to a very large number of user utterances~\cite{generic,generic_emo}. \texttt{DA} and emotion identification~\cite{classif,heavy_tailed} are done through sequence labelling systems that are usually trained on large corpora (with over $100k$ labelled utterances) such as Switchboard~\cite{datase_swda}, \texttt{MRDA}~\cite{dataset_mrda} or Daily Dialog Act~\cite{dataset_dailydialog}.
Even though large corpora enable learning complex models from scratch (e.g., seq2seq~\cite{colombo2020guiding}), those models are very specific to the labelling scheme employed. Adapting them to different sets of emotions or dialog acts would require more annotated data.\\
Generic representations \cite{mikolov2013distributed,pennington2014glove,peters2018deep,bert,xlnet,roberta} have been shown to be an effective way to adapt models across different sets of labels. Those representations are usually trained on large written corpora such as OSCAR~\cite{oscar}, Book Corpus~\cite{book} or Wikipedia~\cite{wiki}. Although achieving state-of-the-art (SOTA) results on written benchmarks~\cite{glue}, they are not tailored to spoken dialog (\texttt{SD}). Indeed, \citet{tran2019role} have suggested that training a parser on conversational speech data can improve results, due to the discrepancy between spoken and written language (e.g., disfluencies~\cite{disfluency}, fillers~\cite{fillers,dinkar2020importance}, different data distribution).
Furthermore, capturing discourse-level features, which distinguish dialog from other types of text~\cite{context}, e.g., capturing multi-utterance dependencies, is key to embed dialog that is not explicitly present in pre-training objectives~\cite{bert,xlnet,roberta}, as they often treat sentences as a simple stream of tokens. \\
The goal of this work is to train on \texttt{SD} data a generic dialog encoder capturing discourse-level features that produce representations adapted to spoken dialog.
We evaluate these representations on both \texttt{DA} and \texttt{E/S} labelling through a new benchmark \texttt{SILICONE} (Sequence labellIng evaLuatIon benChmark fOr spoken laNguagE) composed of datasets of varying sizes using different sets of labels. 
We place ourselves in the general trend of using smaller models to obtain lightweight representations~\cite{tiny,albert} that can be trained without a costly computation infrastructure while achieving good performance on several downstream tasks~\cite{efficient}.
Concretely, since hierarchy is an inherent characteristic of dialog~\cite{context}, we propose the first hierarchical generic multi-utterance encoder based on a hierarchy of transformers. This allows us to factorise the model parameters, getting rid of long term dependencies and enabling training on a reduced number of GPUs.
Based on this hierarchical structure, we generalise two existing pre-training objectives. As embeddings highly depend on data quality~\cite{flaubert} and volume~\cite{roberta}, we preprocess OpenSubtitles~\cite{open}: a large corpus of spoken dialog from movies. This corpora is an order of magnitude bigger than corpora~\cite{budzianowski2018multiwoz,ubuntu,cornell} used in previous works~\cite{mehri2019pretraining,emotion_transfert}. Lastly, we evaluate our encoder along with other baselines on \texttt{SILICONE}, which lets us draw finer conclusions of the generalisation capability of our models\footnote{Upon publication, we will release the code, models and especially the preprocessing scripts to replicate our results.}.

\section{Method}\label{sec:model}
We start by formally defining the Sequence Labelling Problem. At the highest level, we have a set $D$ of conversations composed of utterances, i.e., $D = (C_1,C_2,\dots,C_{|D|})$ with $Y= (Y_1,Y_2,\dots,Y_{|D|})$ being the corresponding set of labels (e.g., \texttt{DA}, \texttt{E/S}). At a lower level each conversation $C_i$ is composed of utterances $u$, i.e $C_i= (u_1,u_2,\dots,u_{|C_i|})$ with $Y_i = (y_1, y_2, \dots, y_{|C_i|})$ being the corresponding sequence of labels: each $u_i$ is associated with a unique label $y_i$. At the lowest level, each utterance $u_i$ can be seen as a sequence of words, i.e $u_i = (\omega^i_1, \omega^i_2, \dots, \omega^i_{|u_i|})$. Concrete examples with dialog act can be found in \autoref{tab:comp_ex}.

\begin{table}[!htb]\label{tab:kysymys}
\centering
\resizebox{.47\textwidth}{!}{\begin{tabular}{l|c} 
\hline
Utterances & \texttt{DA}  \\
\hline
How long does that take you to get to work? &	\texttt{qw}\\
Uh, about forty-five, fifty minutes.&	\texttt{sd} \\
\makecell[l]{How does that work, work out with, uh,\\storing your bike and showering and all that?}
&	\texttt{qw} \\
Yeah , &	\texttt{b}\\
It can be a pain .	&\texttt{sd}\\
\hline
\makecell[l]{It's, it's nice riding to school because\\ it's all along a canal path, uh,}&	\texttt{sd} \\ 
\makecell[l]{Because it's just, \\it's along the Erie Canal up here.}&	\texttt{sd} \\ 
So, what school is it?&	\texttt{qw}  \\ 
Uh, University of Rochester.&	\texttt{sd} \\ 
Oh, okay.& \texttt{bk} \\
\hline
\end{tabular}}
\caption{Examples of dialogs labelled with \texttt{DA} taken from \texttt{SwDA}. The labels  \texttt{qw}, \texttt{sd}, \texttt{b}, \texttt{bk} respectively correspond to wh-question, statement-non-opinion, backchannel and response acknowledgement.}
\label{tab:comp_ex}
\end{table}

\subsection{Pre-training Objectives}\label{ssec:notations}
Our work builds upon existing objectives designed to pre-train encoders: the Masked Language Model (\texttt{MLM}) from~\citet{bert,roberta,albert,zhang2019hibert} and the Generalized Autoregressive Pre-training (\texttt{GAP}) from \citet{xlnet}.

\noindent\textbf{\texttt{MLM} Loss}: The \texttt{MLM} loss corrupts sequences (or in our case, utterances) by masking a proportion $p_\omega$ of tokens. The model learns bidirectional representations by predicting the original identities of the masked-out tokens. Formally, for an utterance $u_i$, a random set of indexed positions $m^{u_i}$ is selected and the associated tokens are replaced by a masked token \texttt{[MASK]} to obtain a corrupted utterance $u^\text{masked}_i$. The set of parameters $\theta$ is learnt by maximizing : 
\begin{equation}
     \mathcal{L}^u_{\texttt{MLM}}(\theta,u_i)=\mathbb{E}\left[\sum_{t \in m^{u_i}} \log(p_{\theta}(\omega^i_t | \Tilde{u}_i))\right]
     \label{eq:mlm_loss}
\end{equation} 
where $\Tilde{u}_i$ is the corrupted utterance, $m^{u_i}_j \sim \textit{unif}\{1, |u_i|\}$ $\forall$ $j \in [1,p_\omega]$ 
and $p_\omega$ is the proportion of masked tokens.

\noindent\textbf{\texttt{GAP} Loss}: the \texttt{GAP} loss consists in computing a classic language modelling loss across different factorisation orders of the tokens. In this way, the model will learn to gather information across all possible positions from both directions. The set of parameters $\theta$ is learnt by maximising:
\begin{equation}\label{eq:gap_loss}
\mathcal{L}^u_{\texttt{GAP}}(\theta,u_i) = \mathbb{E}\left[\mathbb{E}_{\textbf{z} \sim \mathbb{Z}_{|u_i|}} \bigg[ \sum_{t} \log p_{\theta}(\omega^i_{z_t} | u_i^{{\textbf{z}<t}})\bigg]\right]
\end{equation} where $\mathbb{Z}_{|u_i|}$ is the set of permutations of length $|u_i|$ and $u_i^{{\textbf{z}<t}}$ represent the first $t$ tokens of $u_i$ when permuting the sequence according to $\textbf{z} \in \mathbb{Z}_{|u_i|}$. 

\subsection{Hierarchical Encoding}
Capturing dependencies at different granularity levels is key for dialog embedding.
Thus, we choose a hierarchical encoder~\cite{crf_chen, sota_swda_1}. 
It is composed of two functions $f^u$ and $f^c$, satisfying:
\begin{align}\label{eq:multi_obj}
        \mathcal{E}_{u_i} &= f^u_\theta(\omega_1, \hdots, \omega_{|u_i|}) \\
        \mathcal{E}_{C_j} &= f^d_\theta( \mathcal{E}_{u_1}, \hdots, \mathcal{E}_{C_j})
\end{align}
where $\mathcal{E}_{u_i} \in \mathbb{R}^{d_u}$ is the embedding of $u_i$ and $\mathcal{E}_{C_j} \in \mathbb{R}^{d_d}$ the embedding of $C_j$. The structure of the hierarchical encoder is depicted in~\autoref{fig:pretraining_fig}.

\subsection{Hierarchical Pre-training}
\subsubsection{General Motivation}
\begin{figure}[ht]
    \centering
    \includegraphics[width=0.46\textwidth]{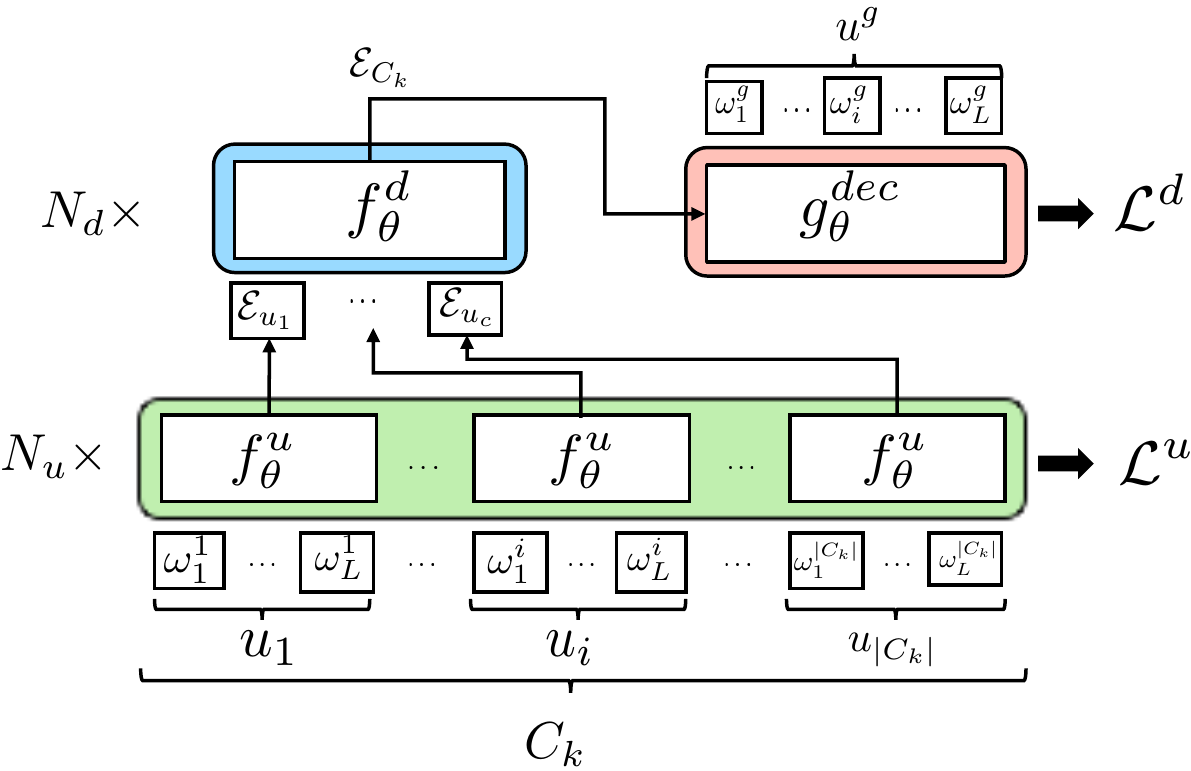}
    \caption{General structure of our proposed hierarchical dialog encoder, with a decoder: $f^u_\theta$, $f^d_\theta$ and the sequence label decoder ($g_\theta^{dec}$) are colored respectively in green, blue and red.}
    \label{fig:pretraining_fig}\vspace{-.3cm}
\end{figure}
Current self-supervised pre-training objectives such as \texttt{MLM} and \texttt{GAP} are trained at the sequence level, which for us translates to only learning $f^u_\theta$.
In this section, we extend both the \texttt{MLM} and \texttt{GAP} losses at the dialog level in order to pre-train $f^d_\theta$.
Following previous work on both multi-task learning~\cite{multi_task_1,multi_task_2} and hierarchical supervision~\cite{garcia2019token,hierarchy_loss}, we argue that optimising simultaneously at both levels rather than separately improves the quality of the resulting embeddings.
Thus, we write our global hierarchical loss as:
\begin{equation}\label{eq:h_loss}
    \mathcal{L}(\theta) = \lambda_u * \mathcal{L}^u(\theta) +  \lambda_d *\mathcal{L}^d(\theta)
\end{equation}
where $\mathcal{L}^u(\theta)$ is either the \texttt{MLM} or \texttt{GAP} loss at the utterance level and $\mathcal{L}^d(\theta)$ is its generalisation at the dialog level.


\subsubsection{\texttt{MLM} Loss}
The \texttt{MLM} loss at the utterance level is defined in~\autoref{eq:mlm_loss}. Our generalisation at the dialog level masks a proportion $p_\mathcal{C}$ of utterances and generates the sequences of masked tokens (a concrete example can be found in \autoref{sec:sup_model}).  
Thus, at the dialog level the \texttt{MLM} loss is defined as: 
\begin{equation}
    \mathcal{L}^d_{\texttt{MLM}}(\theta,C_k)=\mathbb{E}\left[\sum_{j \in m^{C_k}}\sum_{i = 1}^{|u_j|} \log(p_{\theta}(\omega^j_i | \Tilde{C}_k))\right]
\end{equation} 
where $m^{C_k}_j \sim \textit{unif}\{1, |C_k|\}$ $\forall$ $j \in [1,p_\mathcal{C}]$ is the set of positions of masked utterances in the context $C_k$, $\Tilde{C}_k$ is the corrupted context, and $p_{\mathcal{C}}$ is the proportion of masked utterances.  

\subsubsection{\texttt{GAP} Loss} 
The \texttt{GAP} loss at the utterance level is defined in~\autoref{eq:gap_loss}. A possible generalisation of the \texttt{GAP} at the dialog level is to compute the loss of the generated utterance across all factorization orders of the context utterances. Formally, the \texttt{GAP} loss is defined at the dialog level as: 
\begin{align}
\begin{split}
    \mathcal{L}^d_{\texttt{GAP}}(\theta, C_k) &=\\ &\mathbb{E}\left[\mathbb{E}_{\textbf{z} \sim \mathbb{Z}_T} \bigg[ \sum_{t=1}^{| C_k|} \sum_{i=1}^{|u_{z_t}|} \log p_{\theta}(\omega^{z_t}_{i} |  C_k^{\textbf{z}<t})\bigg]\right]
    \end{split}
\end{align}
where  $\omega^{z_t}_i$ denotes the first $i$-th tokens of the permuted $t$-th utterance when permuting the context according to  $\textbf{z} \in \mathbb{Z}_T$ and $C_k^{{\textbf{z}<t}}$ the first $t$ utterances of $C_k$ when permuting the context according to $\textbf{z}$.

\subsection{Architecture}
Commonly, The functions $f^u_\theta$ and $f^d_\theta$ are either modelled with recurrent cells~\cite{hred} or Transformer blocks~\cite{attention_is}. Transformer blocks are more parallelizable, offering shorter paths for the forward and backward signals and requiring significantly less time to train compared to recurrent layers. To the best of our knowledge this is the first attempt to pre-train a hierarchical encoder based only on transformers\footnote{Although it is possible to relax the fixed size imposed by transformers~\cite{dai2019transformer} in this paper we follow~\cite{colombo2020guiding} and fix the context size to $5$ and the max utterance length to $50$ --- these choices are made to work with OpenSubtitles, since the number of available dialogs drops when considering a number of utterances greater than $5$.}. \\
The structure of the model can be found in~\autoref{fig:pretraining_fig}. In order to optimize dialog level losses as described in~\autoref{eq:h_loss}, we generate (through $g_\theta^{dec}$) the sequence with a Transformer Decoder ($\mathcal{T}_{dec}$).
For downstream tasks, the context embedding $\mathcal{E}_{C_k}$ is fed to a simple MLP (simple classification), or to a CRF/GRU/LSTM (sequential prediction) --- see~\autoref{sec:sup_model} for more details. In the rest of the paper, we will name our hierarchical transformer-based encoder $\mathcal{H}\mathcal{T}$ and the hierarchical RNN-based encoder $\mathcal{H}\mathcal{R}$.
We use $\theta^x_y$ to refer to the set of model parameters learnt using the pre-training objective $y$ (either \texttt{MLM} or \texttt{GAP}) at the level $x$\footnote{if $x=u$ solely utterance level training is used, if $x=d$ solely dialog level is used and if $x=u,d$ multi level supervision is used ($\lambda_u,\lambda_d \in \{0,1\}^2$ according to the case.)}.


\subsection{Pre-training Datasets}
Datasets used to pre-train dialog encoders~\cite{emotion_transfert,mehri2019pretraining} are often medium-sized (e.g. Cornell Movie Corpus~\cite{cornell}, Ubuntu~\cite{ubuntu}, MultiWOz~\cite{multiwoz}).
In our work, we focus on OpenSubtitles~\cite{opensub}\footnote{http://opus.nlpl.eu/OpenSubtitles-alt-v2018.php} because (1) it contains spoken language, contrarily to the Ubuntu corpus~\cite{ubuntu} based on logs; (2) as Wizard of Oz~\cite{multiwoz} and Cornell Movie Dialog Corpus~\cite{cornell}, it is a multi-party dataset; and (3) OpenSubtitles is an order of magnitude larger than any other spoken language dataset used in previous work.
We segment OpenSubtitles by considering the duration of the silence between two consecutive utterances. Two consecutive utterances belong to the same conversation if the silence is shorter than $\delta_T$\footnote{We choose $\delta_T=6s$}. Conversations shorter than the context size $T$ are dropped\footnote{Using pre-training method based on the next utterance proposed by \citet{mehri2019pretraining} requires dropping conversation shorter than $T+1$ leading to a non-negligible loss in the preprocessing stage.}. After preprocessing, Opensubtitles contains subtitles from $446520$ movies or series which represent $54642424$ conversations and over $2.3$ billion of words. 

\subsection{Baseline Encoder}
We compare the different methods we presented with two different types of baseline encoders: pre-trained encoders, and hierarchical encoders based on recurrent cells. The latter, achieve current SOTA performance in many sequence labelling tasks~\cite{sota_swda_1,colombo2020guiding,self_attention}.

\noindent\textbf{Pre-trained Encoder Models}.
We use BERT~\cite{bert} through the pytorch implementation provided by the Hugging Face transformers library~\cite{Wolf2019HuggingFacesTS}. The pre-trained model is fed with a concatenation of the utterances. Formally given an input context $C_k = (u_1, \dots u_T)$ the concatenation $[u_1,\dots,u_T]$ is fed to BERT. 

\noindent\textbf{Hierarchical Recurrent Encoders}. In this work we rely on our own implementation of the model based on $\mathcal{H}\mathcal{R}$. Hyperparameters are described in~\autoref{sec:sup_model}.

\section{Evaluation of Sequence Labelling}
\subsection{Related Work}
Sequence labelling tasks for spoken dialog mainly involve two different types of labels: \texttt{DA} and \texttt{E/S}. Early work has tackled the sequence labelling problem as an independent classification of each utterance. Deep neural network models that currently achieve the best results~\cite{bayesian_dialog,svm_dialog,hmm_dialog} model both contextual dependencies between utterances~\cite{colombo2020guiding,concurrent_mrda} and labels~\cite{crf_chen,crf_kumar,crf_li}.

The aforementioned methods require large corpora to train models from scratch, such as: Switchboard Dialog Act (\texttt{SwDA})~\cite{datase_swda},  Meeting Recorder Dialog Act (\texttt{MRDA})~\cite{dataset_mrda}, Daily Dialog Act~\cite{dataset_dailydialog}, HCRC Map Task Corpus~(\texttt{MT})~\cite{dataset_maptask}.
This makes harder their adoption to smaller datasets, such as: Loqui human-human dialogue corpus (\texttt{Loqui})~\cite{dataset_loquihuman}, BT Oasis Corpus (\texttt{Oasis})~\cite{dataset_gtoasis}, Multimodal Multi-Party Dataset (\texttt{MELD})~\cite{dataset_meld}, Interactive emotional dyadic motion capture database (\texttt{IEMO}), SEMAINE database (\texttt{SEM})~\cite{dataset_semaine}. 

\subsection{Presentation of \texttt{SILICONE}}
Despite the similarity between methods usually employed to tackle \texttt{DA} and \texttt{E/S} sequential classification, studies usually rely on a single type of label. Moreover, despite the variety of small or medium-sized labelled datasets, evaluation is usually done on the largest available corpora (e.g., \texttt{SwDA}, \texttt{MRDA}).
We introduce \texttt{SILICONE}, a collection of sequence labelling tasks, gathering both \texttt{DA} and \texttt{E/S} annotated datasets. \texttt{SILICONE} is built upon preexisting datasets which have been considered by the community as challenging and interesting.
Any model that is able to process multiple sequences as inputs and predict the corresponding labels can be evaluated on \texttt{SILICONE}. 
We especially include small-sized datasets, as we believe it will ensure that well-performing models are able to both distil substantial knowledge and adapt to different sets of labels without relying on a large number of examples.
The description of the datasets composing the benchmark can be found in the following sections, while corpora statistics are gathered in~\autoref{tab:presentation_of_silicon}.

\subsubsection{\texttt{DA} Datasets}

\noindent\textbf{Switchboard Dialog Act Corpus (\texttt{SwDA})} is a telephone speech corpus consisting  of two-sided telephone conversations with provided topics. This dataset includes additional features such as speaker id and topic information. The SOTA model, based on a seq2seq architecture with guided attention, reports an accuracy of $85.5\%$~\cite{colombo2020guiding} on the official split.

\noindent\textbf{ICSI MRDA Corpus (\texttt{MRDA})} has been introduced by~\citet{dataset_mrda}. It contains transcripts of multi-party meetings hand-annotated with \texttt{DA}. It is the second biggest dataset with around $110k$ utterances. The SOTA model reaches an accuracy of $92.2\%$ \cite{sota_swda_1} and uses Bi-LSTMs with attention as encoder as well as additional features, such as the topic of the transcript. 

\noindent\textbf{DailyDialog Act Corpus ({$\texttt{DyDA}_\texttt{a}$})} has been produced by~\citet{dataset_dailydialog}. It contains multi-turn dialogues, supposed to reflect daily communication by covering topics about daily life. The dataset is manually labelled with dialog act and emotions. It is the third biggest corpus of \texttt{SILICONE} with $102k$ utterances. The SOTA model reports an accuracy of $88.1\%$~\cite{sota_swda_1}, using Bi-LSTMs with attention as well as additional features.
We follow the official split introduced by the authors.

\noindent\textbf{HCRC MapTask Corpus (\texttt{MT})} has been introduced by~\cite{dataset_maptask}. To build this corpus, participants were asked to collaborate verbally by describing a route from a first participant’s map by using the map of another participant. This corpus is small ($27k$ utterances). As there is no standard train/dev/test split\footnote{We split according to the code in https://github.com/NathanDuran/Maptask-Corpus.} performances depends on the split. \citet{tran2017generative} make use of a Hierarchical LSTM encoder with a GRU decoder layer and achieves an accuracy of $65.9\%$.

\noindent\textbf{Bt Oasis Corpus (\texttt{Oasis})} contains the transcripts of live calls made to the BT and operator services. This corpus has been introduced by~\cite{dataset_gtoasis} and is rather small ($15k$ utterances). There is no standard train/dev/test split \footnote{We use a random split from https://github.com/NathanDuran/BT-Oasis-Corpus.} and few studies use this dataset. 

\subsubsection{\texttt{S/E} Datasets}
In \texttt{S/E} recognition for spoken language, there is no consensus on the choice the evaluation metric (e.g., \citet{weighted_preproc,weighted_no} use a weighted F-score while \citet{accuracy_fscore} report accuracy). For \texttt{SILICONE}, we choose to stay consistent with the \texttt{DA} research and thus follow~\citet{accuracy_fscore} by reporting the accuracy. Additionally, emotion/sentiment labels are neither merged nor prepossessed\footnote{Comparison with concurrent work is more difficult as system performance heavily depends on the number of classes and label processing varies across studies \cite{variability}.}.

\noindent\textbf{DailyDialog Emotion Corpus ($\texttt{DyDA}_\texttt{e}$)} has been previously introduced and contains eleven emotional labels. The SOTA model~\cite{de2019joint} is based on BERT with additional Valence Arousal and Dominance features and reaches an accuracy of 85\% on the official split.

\noindent\textbf{Multimodal EmotionLines Dataset (\texttt{MELD})} has been created by enhancing and extending EmotionLines dataset~\cite{emotion_lines} where multiple speakers participated in the dialogues. There are two types of annotations $\texttt{MELD}_\texttt{s}$ and  $\texttt{MELD}_\texttt{e}$: three sentiments (positive, negative and neutral) and seven emotions (anger, disgust, fear, joy,neutral, sadness and surprise). The SOTA model with text only is proposed by \citet{accuracy_fscore} and is inspired by quantum physics. On the official split, it is compared with a hierarchical bi-LSTM, which it beats with an accuracy of $61.9$\% ($\texttt{MELD}_\texttt{s}$) and $67.9$\% ($\texttt{MELD}_\texttt{e}$) against $60.8\%$ and $65.2$.

\noindent\textbf{IEMOCAP database (\texttt{IEMO})} is a multimodal database of ten speakers. It consists of dyadic sessions where actors perform improvisations or scripted scenarios. Emotion categories are: anger, happiness, sadness, neutral, excitement, frustration, fear, surprise, and other. There is no official split on this dataset. One proposed model is built with bi-LSTMs and achieves $35.1\%$, with text only \cite{accuracy_fscore}.

\noindent\textbf{SEMAINE database (\texttt{SEM})} comes from the Sustained Emotionally coloured Machine human Interaction using Nonverbal Expression project~\cite{dataset_semaine}. This dataset has been annotated on three sentiments labels: positive, negative and neutral by \citet{barriere2018attitude}. It is built on Multimodal Wizard of Oz experiment where participants held conversations with an operator who adopted various roles designed to evoke emotional reactions. There is no official split on this dataset.
\begin{table*}
\begin{center}
\resizebox{.7\textwidth}{!}{\begin{tabular}{c  c c c c c c c} 
 \hline
 Corpus  & $\lvert Train \rvert$ & $\lvert Val \rvert$ & $\lvert Test \rvert$ & Utt. & $\lvert Labels \rvert$ & Task& Utt./$\lvert Labels \rvert$ \\ 
  \hline
  $\texttt{SwDA}^\star$  &  1k  & 100 & 11 & 200k   & 42 & \texttt{DA} & 4.8k\\
  $\texttt{MRDA}^\star$ &  56 & 6 &  12 &110k& 5& \texttt{DA} & 2.6k\\
      $\texttt{DyDA}_\texttt{a}$&  11k  & 1k & 1k & 102k  & 4 & \texttt{DA}& 25.5k \\
  $\texttt{MT}^\star$ &  121 & 22 & 25 &  36k & 12& \texttt{DA} & 3k\\
  $\texttt{Oasis}^\star$ &  508 & 64 & 64 & 15k & 42 & \texttt{DA} & 357\\
 \hline
    $\texttt{DyDA}_\texttt{e}$ &   11k  & 1k & 1k & 102k  & 7 &\texttt{E}&2.2k \\
    $\texttt{MELD}_\texttt{s}^\star$ &  934 & 104  &  280 & 13k  & 3 & \texttt{S} & 4.3k\\
        $\texttt{MELD}_\texttt{e}^\star$& 934 & 104  &  280 & 13k  & 7 & \texttt{S} & 1.8k\\
    \texttt{IEMO}&  108 & 12 & 31 & 10k & 6 & \texttt{E}&1.7k\\
      \texttt{SEM}  & 62 & 7 & 10 & 5,6k & 3 & \texttt{S}& 1.9k\\
\end{tabular}}
\end{center}
\caption{Statistics of datasets composing  \texttt{SILICONE}. \texttt{E} stands for emotion label and \texttt{S} for sentiment label; $^\star$ stands for datasets with available official split. Sizes of Train, Val and Test are given in number of conversations. }\label{tab:presentation_of_silicon}
\end{table*}

\section{Results on \texttt{SILICONE}}
This section gathers experiments performed on the \texttt{SILICONE} benchmark. 
We first analyse an appropriate choice for the decoder, which is selected over a set of experiments on our baseline encoders: a pre-trained BERT model and a hierarchical RNN-based encoder ($\mathcal{H}\mathcal{R}$).
Since we focus on small-sized pre-trained representations, we limit the sizes of our pre-trained models to \texttt{TINY} and \texttt{SMALL} (see \autoref{tab:model_size}). We then study the results of the baselines and our hierarchical transformer encoders ($\mathcal{H}\mathcal{T}$) on \texttt{SILICONE} along three axes: the accuracy of the models, the difference in performance between the \texttt{E/S} and the \texttt{DA} corpora, and the importance of pre-training.
As we aim to obtain robust representations, we do not perform an exhaustive grid search on the downstream tasks.
\subsection{Decoder Choice}
Current research efforts focus on single label prediction, as it seems to be a natural choice for sequence labelling problems (\autoref{ssec:notations}).
Sequence labelling is usually performed with CRFs \cite{crf_chen, crf_kumar} and GRU decoding \cite{colombo2020guiding}, however, it is not clear to what extent inter-label dependencies are already captured by the contextualised encoders, and whether a plain MLP decoder could achieve competitive results.
As can be seen in \autoref{tab:baseline}, we found that in the case of \texttt{E/S} prediction there is no clear difference between CRFs and MLPs, while GRU decoders exhibit poor performance, probably due to a lack of training data.
It is also important to notice, that training a sequential decoder usually requires thorough hyper-parameter fine-tuning.
As our goal is to learn and evaluate general representations that are decoder agnostic, in the following, we will use a plain MLP decoder for all the models compared.
\begin{table}[]
\begin{center}
 \resizebox{.4\textwidth}{!}{\begin{tabular}{ c |c c c } 
 \hline
 &Avg  & Avg \texttt{DA} &   Avg \texttt{E/S} \\ \hline
BERT (+MLP) & 72,8& 81.5  & 64.0 \\
BERT (+GRU) &69.9& 80.4 & 59.3 \\ 
BERT (+CRF) &72.8&81.5 & 64.1   \\ \hline
$\mathcal{H}\mathcal{R}$ (+MLP)  &69.8&  79.1 & 60.4   \\
$\mathcal{H}\mathcal{R}$ (+GRU) &67.6& 79.4& 55.7   \\
$\mathcal{H}\mathcal{R}$ (+CRF) & 70.5&80.3 & 60.7   \\
\end{tabular}}
\caption{Experiments comparing decoder performances. Results are given on \texttt{SILICONE} for two types of baseline encoders (pre-trained BERT models and hierarchical recurrent encoders $\mathcal{H}\mathcal{R}$).}\label{tab:baseline}
\end{center}
\end{table}
\begin{table*}
\setlength{\tabcolsep}{4pt}
\begin{center}
 \resizebox{\textwidth}{!}{
 \begin{tabular}{c || c || c c c  c c | c c c c c } 
 \hline
 & \textbf{Avg} & \texttt{SwDA}  & \texttt{MRDA} & $\texttt{DyDA}_{\texttt{DA}}$ &$\texttt{MT}$ & $\texttt{Oasis}$  &  $\texttt{DyDA}_\texttt{e}$  & $\texttt{MELD}_s$  & $\texttt{MELD}_\texttt{e}$ & $\texttt{IEMO}$ & $\texttt{SEM}$ \\ \hline
    BERT-4layers& 70.4 & 77.8 & 90.7  & 79.0 &88.4 & 66.8 & 90.3 & 55.3  &53.4 & 43.0 &58.8\\
    BERT& 72.8 & 79.2  &90.7  & \textbf{82.6} & 88.2   &66.9& 91.9 & 59.3  &\textbf{61.4} & \textbf{45.0} &62.7\\
     $\mathcal{H}\mathcal{R}$& 69.8 & 77,5  &90,9 & 80,1 & 82,8 & 64,3 & 91.5 &59,3 & 59.9 & 40.3& 51.1 \\
     $\mathcal{H}\mathcal{T}(\theta_{MLM}^{u,d})_{\texttt{(TINY)}}$ &73.3& \textbf{79.3}  &92.0& 80.1 & 90.0   &68,3  & 92.5 &62.6 & 59.9 &42.0&66.6\\
     $\mathcal{H}\mathcal{T}(\theta_{GAP}^{d})_{\texttt{(TINY)}}$ & 71.6 & 78.6  & 91.8 & 78.1 & 89.3   & 64.1 & 91.6 & 60.5 & 55.7 & 42.2 & 63.9 \\
    $\mathcal{H}\mathcal{T}(\theta_{MLM}^{u,d})_{\texttt{(SMALL)}}$ & \textbf{74.3} & 79.2  & \textbf{92.4}& 81.5 & \textbf{90.6}   &\textbf{69.4}  & \textbf{92.7} & \textbf{64.1} &60.1& \textbf{45.0} & \textbf{68.2}
\end{tabular}
}
\caption{Performances of different encoders when decoding using a MLP on \texttt{SILICONE}. The datasets are grouped by label type (\texttt{DA} vs \texttt{E/S}) and ordered by decreasing size. $\texttt{MT}$ stands for $\texttt{Map Task}$, $\texttt{IEM}$ for $\texttt{IEMOCAP}$ and $\texttt{Sem}$ for $\texttt{Semaine}$.}
\label{tab:results}
\end{center}
\end{table*}

\subsection{General Performance Analysis}
\autoref{tab:results} provides an exhaustive comparison of the different encoders over the \texttt{SILICONE} benchmark.
As previously discussed, we adopt a plain MLP as a decoder to compare the different encoders.
We show that \texttt{SILICONE} covers a set of challenging tasks as the best performing model achieves an average accuracy of $74.3$.
Moreover, we observe that despite having half the parameters of a BERT model, our proposed model achieves an average result that is $2\%$ higher on the benchmark.
\texttt{SILICONE} covers two different sequence labelling tasks: \texttt{DA} and \texttt{E/S}.
In \autoref{tab:results} and \autoref{tab:baseline}, we can see that all models exhibit a consistently higher average accuracy (up to $14\%$) on \texttt{DA} tagging compared to \texttt{E/S} prediction. This performance drop could be explained by the different sizes of the corpora (see \autoref{tab:presentation_of_silicon}). 
Despite having a larger number of utterances per label ($u/l$), \texttt{E/S} tasks seem generally harder to tackle for the models. For example, on \texttt{Oasis}, where the $u/l$ is inferior than those of most \texttt{E/S} datasets ($\texttt{MELD}_\texttt{s}$, $\texttt{MELD}_\texttt{e}$, \texttt{IEMO} and \texttt{SEM}), models consistently achieve better results.
 
\subsection{Importance of Pre-training for \texttt{SILICONE}} 
\begin{figure}
  \centering
  \includegraphics[width=0.45\textwidth]{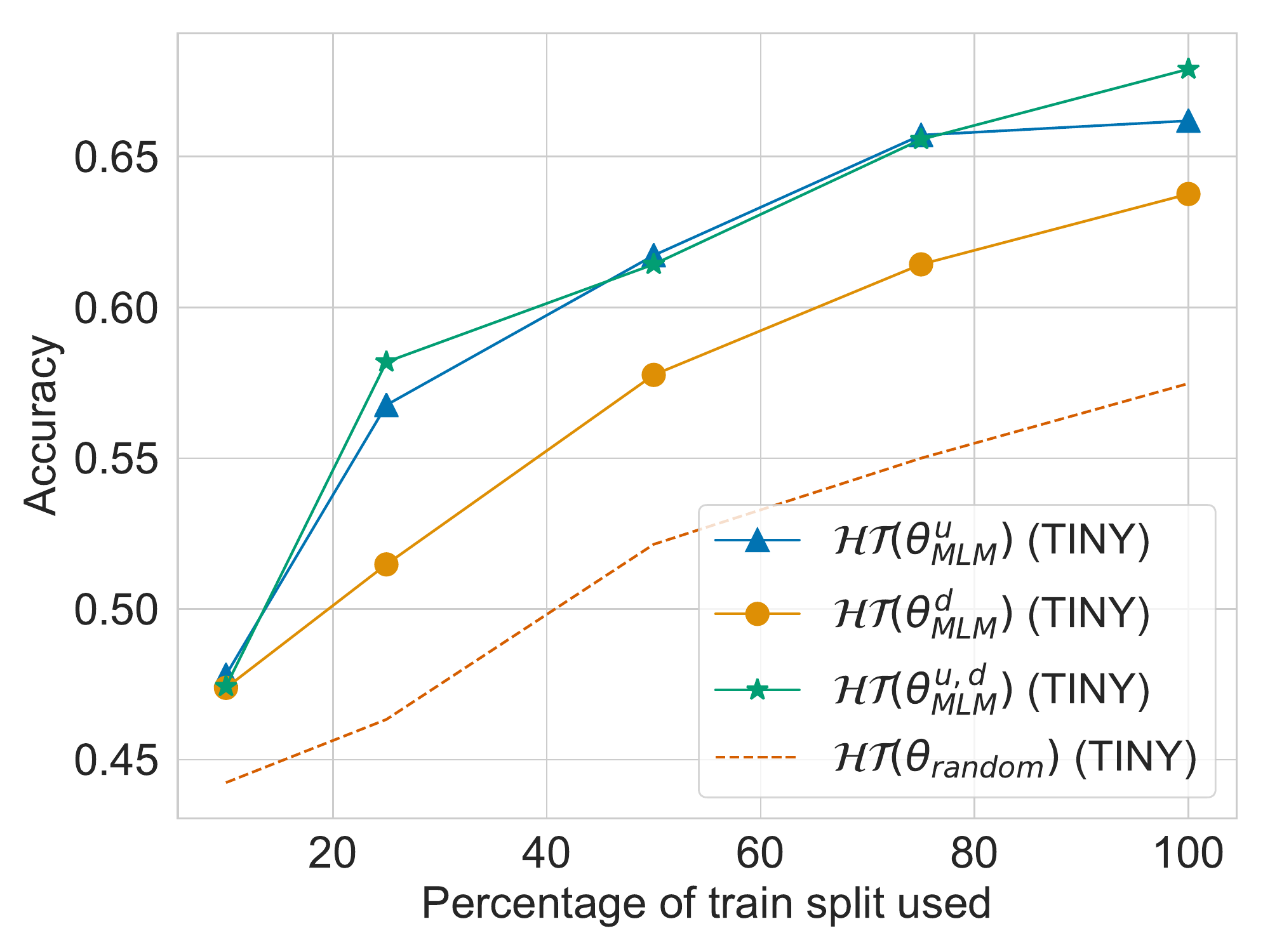}
    \caption{A comparison of pre-trained encoders being fine-tuned on different percentage the training set of \texttt{SEM}. Validation and test set are fixed over all experiments, reported scores are averaged over 10 different random split.}
    \label{fig:exp_split_semaine}
\end{figure}

Results reported in \autoref{tab:results} and \autoref{tab:baseline} show that pre-trained transformer-based encoders achieve consistently higher accuracy on \texttt{SILICONE}, even when they are not explicitly considering the hierarchical structure.
This difference can be observed both in small-sized datasets (e.g. \texttt{MELD} and \texttt{SEM}) and in medium/large size datasets (e.g \texttt{SwDA} and \texttt{MRDA}).
To validate the importance of pre-training in a regime of low data, we train different $\mathcal{H}\mathcal{T}$ (with random initialisation) on different portions of \texttt{SEM} and $\texttt{MELD}_\texttt{s}$.
Results shown in \autoref{fig:exp_split_semaine} illustrate the importance of pre-trained representations.
\section{Model Analysis}
In this section, we dissect our hierarchical pre-trained models in order to better understand the relative importance of each component. We show how a hierarchical encoder allows us to obtain a light and efficient model. Additional experiments can be found in \autoref{supp:results}.
\subsection{Pre-training on Spoken vs Written Data}
First, we explore the differences in training representations on spoken and written corpora. 
Experimentally, we compare the predictions on \texttt{SILICONE} made by $\mathcal{H}\mathcal{T}(\theta_{MLM}^u)$ and the one made by $\mathcal{H}\mathcal{T}$($\theta_{BERT-2layers}$).
The latter is a hierarchical encoder where utterance embeddings are obtained with the hidden vector representing the first token [CLS] (see \cite{bert}) of the second layer of BERT.
In both cases, predictions are performed using an MLP\footnote{We consider the two first layer for a fair comparison based on the number of model parameters.}.
Results in \autoref{tab:spoken_language} show higher accuracy when the pre-training is performed on spoken data.
Since \texttt{SILICONE} is a spoken language benchmark, this result might be due to the specific features of colloquial speech (e.g. disfluencies, sentence length, vocabulary, word frequencies).
\begin{table}
\begin{center}
 \begin{tabular}{ c | c c } 
 \hline
 &Avg \texttt{DA} & Avg \texttt{E/S} \\ \hline
 BERT (4 layers)&80.5 & 60.2  \\
$\mathcal{H}\mathcal{T}$($\theta_{BERT-2layers}$) & 80.5& 61.1\\
$\mathcal{H}\mathcal{T}$ ($\theta_{MLM}^u$) &\textbf{80.8} & \textbf{64.0}\\ 
\end{tabular}
\caption{Results of ablation studies on \texttt{SILICONE}}
 \label{tab:spoken_language}
\end{center}
\end{table}
\subsection{Hierarchy and Multi-Level Supervision}
We study the relative importance of three aspects of our hierarchical pre-training with multi-level supervision. 
We first show that accounting for the hierarchy increases the performance of fine-tuned encoders, even without our specific pre-training procedure.
We then compare our two proposed hierarchical pre-training procedures based on the \texttt{GAP} or \texttt{MLM} loss. Lastly, we look at the contribution of the possible levels of supervision on reduced training data from \texttt{SEM}.
\subsubsection{Importance of hierarchical fine-tuning}
We compare the performance of BERT-4layers with the $\mathcal{H}\mathcal{T}$($\theta_{BERT-2layer}$) previously described.
Results reported in \autoref{tab:spoken_language} demonstrate that fine-tuning on downstream tasks with a hierarchical encoder yields to higher accuracy, with fewer parameters, even when using already pre-trained representations.
\subsubsection{\texttt{MLM} vs \texttt{GAP}} 
In this experiment, we compare the different pre-training objectives at utterance and dialog level.
As a reminder $\mathcal{H}\mathcal{T}$($\theta_{MLM}^u$) and $\mathcal{H}\mathcal{T}$($\theta_{GAP}^u$) are respectively trained using the standard \texttt{MLM} loss~\cite{bert} and the standard \texttt{GAP} loss~\cite{xlnet}.
In \autoref{tab:pretraining_comparizon} we report the different pre-training objective results.
We observe that pre-training at the dialog level achieves comparable results to the utterance level pre-training for \texttt{MLM} and slightly worse for \texttt{GAP}.
Interestingly, we observe that $\mathcal{H}\mathcal{T}$($\theta_{GAP}^u$) compared to  $\mathcal{H}\mathcal{T}$($\theta_{MLM}^u$) achieves worse results, which is not consistent with the performance observed on other benchmarks, such as GLUE \cite{glue}.
The lower accuracy of the models trained using a \texttt{GAP}-based loss could be due to several factors (e.g., model size, pre-training using the \texttt{GAP} loss could require a finer choice of hyper-parameters).
Finally, we see that supervising at both dialog and utterance level helps for \texttt{MLM}\footnote{We investigate a similar setting for \texttt{GAP} which lead to poor results, the loss hit a plateau suggesting that objectives are competing against each other. More advanced optimisations techniques \cite{multi_loss_opt} are left for future work.}.
\subsubsection{Multi level Supervision for pre-training}\label{sssec:multi_sup}
In this section, we illustrate the advantages of learning using several levels of supervision on small datasets. We fine-tune different model on \texttt{SEM} using different size of the training set. Results are shown in \autoref{fig:exp_split_semaine}. Overall we see that introducing sequence level supervision induces a consistent improvement on \texttt{SEM}. Results on $\texttt{MELD}_\texttt{s}$ are provided in \autoref{supp:results}.
\begin{table}[]
\begin{center}
 \begin{tabular}{ c | c c } 
 \hline
 & Avg \texttt{DA} &   Avg \texttt{E/S} \\ \hline
$\mathcal{H}\mathcal{T}(\theta_{MLM}^u)$ & 80.8 & 64.0 \\
$\mathcal{H}\mathcal{T}(\theta_{MLM}^d)$ & 80.8 & 64.0 \\\hline 
$\mathcal{H}\mathcal{T}(\theta_{GAP}^u)$ & 80.7 & 62.0 \\
$\mathcal{H}\mathcal{T}(\theta_{GAP}^d)$ & 80.4 & 62.8 \\\hline  
$\mathcal{H}\mathcal{T}(\theta_{MLM}^{u,d})$ & \textbf{81.9} & \textbf{64.7} \\
\end{tabular}
\caption{Comparison of \texttt{GAP} and \texttt{MLM} with a comparable number of parameters. For all models a MLP decoder is used on top of a \texttt{TINY} pre-trained encoder.}
\label{tab:pretraining_comparizon}
\end{center}
\end{table} 
\subsection{Other advantages of hierarchy}
\begin{table}
\begin{center}
\begin{tabular}{ c |c c c c} 
 \hline
 & Emb. & Word &Seq & Total  \\ \hline
  
    BERT & \multirow{5}{*}{23}  & \multicolumn{2}{c}{87}  & 110 \\
        BERT (4-layer) &  & \multicolumn{2}{c}{43}  & 66 \\
    HMLP &   &8.6 & 7.8  & 40  \\
    (\texttt{TINY}) &   &2.9 & 2.8 & 28.7 \\
     (\texttt{SMALL}) &   &10.6 & 10.6 & 45  \\
\end{tabular}
\caption{Number of parameters for the encoders. Sizes are given in million of parameters.}
\label{tab:model_size}
\end{center}
\end{table}
Introducing a hierarchical design in the encoder allows to break dialog into utterances and to consider inputs of size $T$ instead of size $512$. First, it allows parameters sharing, reducing the number of model parameters. The different model sizes are reported in \autoref{tab:model_size}.
Our \texttt{TINY} model contains half the parameters of BERT (4-layers). Furthermore, modelling long-range dependencies hierarchically makes learning faster and allows to get rid of learning tricks (e.g., partial order prediction~\cite{xlnet}, two-stage pre-training based on sequence length~\cite{bert}) required for non-hierarchical encoders.
Lastly, original BERT and XLNET are pre-trained using respectively 16 and 512 TPUs. Pre-training lasts several days with over $500K$ iterations. Our \texttt{TINY} hierarchical models are pre-trained during $180K$ iterations (1.5 days) on 4 NVIDIA V100.
\section{Conclusions}
In this paper, we propose a hierarchical transformer-based encoder tailored for spoken dialog. We extend two well-known pre-training objectives to adapt them to a hierarchical setting and use OpenSubtitles, the largest spoken language dataset available, for encoder pre-training. Additionally, we provide an evaluation benchmark dedicated to comparing sequence labelling systems for the NLP community, \texttt{SILICONE}, on which we compare our models and pre-training procedures with previous approaches.
By conducting ablation studies, we demonstrate the importance of using a hierarchical structure for the encoder, both for pre-training and fine-tuning.
Finally, we find that our approach is a powerful method to learn generic representations on spoken dialog, with less parameters than state-of-the-art transformer models. 

These results open new future research directions: (1) to investigate new pre-training objectives leveraging the hierarchical framework in order to achieve better results on \texttt{SILICONE} while keeping light models (2) to provide multilingual models using the whole pre-training corpus (OpenSubtitles) available in 62 languages, (3) investigate robust methods \cite{robust_staerman} and the application of our embedding to different anomaly detection settings \cite{anomaly_1,anomaly_2}.
We hope that the \texttt{SILICONE} benchmark, experimental results, and publicly available code encourage further research to build stronger sequence labelling systems for NLP.

\section*{Acknowledgement}
This work was supported by a grant overseen from the French National Research Agency (ANR-17-MAOI).

\newpage
\bibliography{emnlp2020}

\begin{thebibliography}{77}
\expandafter\ifx\csname natexlab\endcsname\relax\def\natexlab#1{#1}\fi

\bibitem[{Agarap(2018)}]{relu}
Abien~Fred Agarap. 2018.
\newblock Deep learning using rectified linear units (relu).
\newblock \emph{arXiv preprint arXiv:1803.08375}.

\bibitem[{Argyriou et~al.(2007)Argyriou, Evgeniou, and Pontil}]{multi_task_1}
Andreas Argyriou, Theodoros Evgeniou, and Massimiliano Pontil. 2007.
\newblock Multi-task feature learning.
\newblock In \emph{Advances in neural information processing systems}, pages
  41--48.

\bibitem[{Barriere et~al.(2018)Barriere, Clavel, and
  Essid}]{barriere2018attitude}
Valentin Barriere, Chlo{\'e} Clavel, and Slim Essid. 2018.
\newblock Attitude classification in adjacency pairs of a human-agent
  interaction with hidden conditional random fields.
\newblock In \emph{2018 IEEE International Conference on Acoustics, Speech and
  Signal Processing (ICASSP)}, pages 4949--4953. IEEE.

\bibitem[{Budzianowski et~al.(2018{\natexlab{a}})Budzianowski, Wen, Tseng,
  Casanueva, Stefan, Osman, and Ga{\v{s}}i\'c}]{multiwoz}
Pawe{\l} Budzianowski, Tsung-Hsien Wen, Bo-Hsiang Tseng, I{\~n}igo Casanueva,
  Ultes Stefan, Ramadan Osman, and Milica Ga{\v{s}}i\'c. 2018{\natexlab{a}}.
\newblock Multiwoz - a large-scale multi-domain wizard-of-oz dataset for
  task-oriented dialogue modelling.
\newblock In \emph{Proceedings of the 2018 Conference on Empirical Methods in
  Natural Language Processing (EMNLP)}.

\bibitem[{Budzianowski et~al.(2018{\natexlab{b}})Budzianowski, Wen, Tseng,
  Casanueva, Ultes, Ramadan, and Ga{\v{s}}i{\'c}}]{budzianowski2018multiwoz}
Pawe{\l} Budzianowski, Tsung-Hsien Wen, Bo-Hsiang Tseng, Inigo Casanueva,
  Stefan Ultes, Osman Ramadan, and Milica Ga{\v{s}}i{\'c}. 2018{\natexlab{b}}.
\newblock Multiwoz-a large-scale multi-domain wizard-of-oz dataset for
  task-oriented dialogue modelling.
\newblock \emph{arXiv preprint arXiv:1810.00278}.

\bibitem[{Chen et~al.(2018{\natexlab{a}})Chen, Hsu, Kuo, Ku
  et~al.}]{emotion_lines}
Sheng-Yeh Chen, Chao-Chun Hsu, Chuan-Chun Kuo, Lun-Wei Ku, et~al.
  2018{\natexlab{a}}.
\newblock Emotionlines: An emotion corpus of multi-party conversations.
\newblock \emph{arXiv preprint arXiv:1802.08379}.

\bibitem[{Chen et~al.(2018{\natexlab{b}})Chen, Yang, Zhao, Cai, and
  He}]{crf_chen}
Zheqian Chen, Rongqin Yang, Zhou Zhao, Deng Cai, and Xiaofei He.
  2018{\natexlab{b}}.
\newblock Dialogue act recognition via crf-attentive structured network.
\newblock In \emph{The 41st International ACM SIGIR Conference on Research \&
  Development in Information Retrieval}, pages 225--234.

\bibitem[{Clavel and Callejas(2015)}]{variability}
Chloe Clavel and Zoraida Callejas. 2015.
\newblock Sentiment analysis: from opinion mining to human-agent interaction.
\newblock \emph{IEEE Transactions on affective computing}, 7(1):74--93.

\bibitem[{Colombo et~al.(2020)Colombo, Chapuis, Manica, Vignon, Varni, and
  Clavel}]{colombo2020guiding}
Pierre Colombo, Emile Chapuis, Matteo Manica, Emmanuel Vignon, Giovanna Varni,
  and Chloe Clavel. 2020.
\newblock Guiding attention in sequence-to-sequence models for dialogue act
  prediction.
\newblock \emph{arXiv preprint arXiv:2002.08801}.

\bibitem[{Colombo et~al.(2019)Colombo, Witon, Modi, Kennedy, and
  Kapadia}]{generic_emo}
Pierre Colombo, Wojciech Witon, Ashutosh Modi, James Kennedy, and Mubbasir
  Kapadia. 2019.
\newblock Affect-driven dialog generation.
\newblock \emph{arXiv preprint arXiv:1904.02793}.

\bibitem[{Dai et~al.(2019)Dai, Yang, Yang, Carbonell, Le, and
  Salakhutdinov}]{dai2019transformer}
Zihang Dai, Zhilin Yang, Yiming Yang, Jaime Carbonell, Quoc~V Le, and Ruslan
  Salakhutdinov. 2019.
\newblock Transformer-xl: Attentive language models beyond a fixed-length
  context.
\newblock \emph{arXiv preprint arXiv:1901.02860}.

\bibitem[{Danescu-Niculescu-Mizil and Lee(2011)}]{cornell}
Cristian Danescu-Niculescu-Mizil and Lillian Lee. 2011.
\newblock Chameleons in imagined conversations: A new approach to understanding
  coordination of linguistic style in dialogs.
\newblock In \emph{Proceedings of the Workshop on Cognitive Modeling and
  Computational Linguistics, ACL 2011}.

\bibitem[{De~Bruyne et~al.(2019)De~Bruyne, Atanasova, and
  Augenstein}]{de2019joint}
Luna De~Bruyne, Pepa Atanasova, and Isabelle Augenstein. 2019.
\newblock Joint emotion label space modelling for affect lexica.
\newblock \emph{arXiv preprint arXiv:1911.08782}.

\bibitem[{Denoyer and Gallinari(2006)}]{wiki}
Ludovic Denoyer and Patrick Gallinari. 2006.
\newblock The wikipedia xml corpus.
\newblock In \emph{International Workshop of the Initiative for the Evaluation
  of XML Retrieval}, pages 12--19. Springer.

\bibitem[{Devlin et~al.(2018)Devlin, Chang, Lee, and Toutanova}]{bert}
Jacob Devlin, Ming-Wei Chang, Kenton Lee, and Kristina Toutanova. 2018.
\newblock Bert: Pre-training of deep bidirectional transformers for language
  understanding.
\newblock \emph{arXiv preprint arXiv:1810.04805}.

\bibitem[{Dinkar et~al.(2020)Dinkar, Colombo, Labeau, and
  Clavel}]{dinkar2020importance}
Tanvi Dinkar, Pierre Colombo, Matthieu Labeau, and Chlo{\'e} Clavel. 2020.
\newblock The importance of fillers for text representations of speech
  transcripts.
\newblock \emph{arXiv preprint arXiv:2009.11340}.

\bibitem[{Garcia et~al.(2019)Garcia, Colombo, Essid, d'Alch{\'e} Buc, and
  Clavel}]{garcia2019token}
Alexandre Garcia, Pierre Colombo, Slim Essid, Florence d'Alch{\'e} Buc, and
  Chlo{\'e} Clavel. 2019.
\newblock From the token to the review: A hierarchical multimodal approach to
  opinion mining.
\newblock \emph{arXiv preprint arXiv:1908.11216}.

\bibitem[{Ghosal et~al.(2019)Ghosal, Majumder, Poria, Chhaya, and
  Gelbukh}]{weighted_preproc}
Deepanway Ghosal, Navonil Majumder, Soujanya Poria, Niyati Chhaya, and
  Alexander Gelbukh. 2019.
\newblock Dialoguegcn: A graph convolutional neural network for emotion
  recognition in conversation.
\newblock \emph{arXiv preprint arXiv:1908.11540}.

\bibitem[{Godfrey et~al.(1992)Godfrey, Holliman, and McDaniel}]{datase_swda}
John~J. Godfrey, Edward~C. Holliman, and Jane McDaniel. 1992.
\newblock Switchboard: Telephone speech corpus for research and development.
\newblock In \emph{Proceedings of the 1992 IEEE International Conference on
  Acoustics, Speech and Signal Processing - Volume 1}, ICASSP’92, page
  517–520, USA. IEEE Computer Society.

\bibitem[{Hazarika et~al.(2019)Hazarika, Poria, Zimmermann, and
  Mihalcea}]{emotion_transfert}
Devamanyu Hazarika, Soujanya Poria, Roger Zimmermann, and Rada Mihalcea. 2019.
\newblock Emotion recognition in conversations with transfer learning from
  generative conversation modeling.
\newblock \emph{arXiv preprint arXiv:1910.04980}.

\bibitem[{Henderson et~al.(2020)Henderson, Hu, Romoff, Brunskill, Jurafsky, and
  Pineau}]{efficient}
Peter Henderson, Jieru Hu, Joshua Romoff, Emma Brunskill, Dan Jurafsky, and
  Joelle Pineau. 2020.
\newblock Towards the systematic reporting of the energy and carbon footprints
  of machine learning.
\newblock \emph{arXiv preprint arXiv:2002.05651}.

\bibitem[{Hendrycks and Gimpel(2016)}]{gelu}
Dan Hendrycks and Kevin Gimpel. 2016.
\newblock Gaussian error linear units (gelus).
\newblock \emph{arXiv preprint arXiv:1606.08415}.

\bibitem[{Jalalzai et~al.(2020)Jalalzai, Colombo, Clavel, Gaussier, Varni,
  Vignon, and Sabourin}]{heavy_tailed}
Hamid Jalalzai, Pierre Colombo, Chlo{\'e} Clavel, Eric Gaussier, Giovanna
  Varni, Emmanuel Vignon, and Anne Sabourin. 2020.
\newblock Heavy-tailed representations, text polarity classification \& data
  augmentation.
\newblock \emph{arXiv preprint arXiv:2003.11593}.

\bibitem[{Jiao et~al.(2019)Jiao, Yin, Shang, Jiang, Chen, Li, Wang, and
  Liu}]{tiny}
Xiaoqi Jiao, Yichun Yin, Lifeng Shang, Xin Jiang, Xiao Chen, Linlin Li, Fang
  Wang, and Qun Liu. 2019.
\newblock Tinybert: Distilling bert for natural language understanding.
\newblock \emph{arXiv preprint arXiv:1909.10351}.

\bibitem[{Keizer et~al.(2002)Keizer, op~den Akker, and
  Nijholt}]{bayesian_dialog}
Simon Keizer, Rieks op~den Akker, and Anton Nijholt. 2002.
\newblock Dialogue act recognition with bayesian networks for dutch dialogues.
\newblock In \emph{Proceedings of the Third SIGdial Workshop on Discourse and
  Dialogue}.

\bibitem[{Kingma and Ba(2014)}]{adam}
Diederik~P Kingma and Jimmy Ba. 2014.
\newblock Adam: A method for stochastic optimization.
\newblock \emph{arXiv preprint arXiv:1412.6980}.

\bibitem[{Kumar et~al.(2018)Kumar, Agarwal, Dasgupta, and Joshi}]{crf_kumar}
Harshit Kumar, Arvind Agarwal, Riddhiman Dasgupta, and Sachindra Joshi. 2018.
\newblock Dialogue act sequence labeling using hierarchical encoder with crf.
\newblock In \emph{Thirty-Second AAAI Conference on Artificial Intelligence}.

\bibitem[{Lan et~al.(2019)Lan, Chen, Goodman, Gimpel, Sharma, and
  Soricut}]{albert}
Zhenzhong Lan, Mingda Chen, Sebastian Goodman, Kevin Gimpel, Piyush Sharma, and
  Radu Soricut. 2019.
\newblock Albert: A lite bert for self-supervised learning of language
  representations.
\newblock \emph{arXiv preprint arXiv:1909.11942}.

\bibitem[{Le et~al.(2019)Le, Vial, Frej, Segonne, Coavoux, Lecouteux, Allauzen,
  Crabb{\'e}, Besacier, and Schwab}]{flaubert}
Hang Le, Lo{\"\i}c Vial, Jibril Frej, Vincent Segonne, Maximin Coavoux,
  Benjamin Lecouteux, Alexandre Allauzen, Beno{\^\i}t Crabb{\'e}, Laurent
  Besacier, and Didier Schwab. 2019.
\newblock Flaubert: Unsupervised language model pre-training for french.
\newblock \emph{arXiv preprint arXiv:1912.05372}.

\bibitem[{Leech and Weisser(2003)}]{dataset_gtoasis}
Geoffrey Leech and Martin Weisser. 2003.
\newblock Generic speech act annotation for task-oriented dialogues.

\bibitem[{Li et~al.(2018{\natexlab{a}})Li, Lin, Collinson, Li, and
  Chen}]{sota_swda_1}
Ruizhe Li, Chenghua Lin, Matthew Collinson, Xiao Li, and Guanyi Chen.
  2018{\natexlab{a}}.
\newblock \href {http://arxiv.org/abs/1810.09154} {A dual-attention
  hierarchical recurrent neural network for dialogue act classification}.
\newblock \emph{CoRR}, abs/1810.09154.

\bibitem[{Li et~al.(2018{\natexlab{b}})Li, Lin, Collinson, Li, and
  Chen}]{concurrent_mrda}
Ruizhe Li, Chenghua Lin, Matthew Collinson, Xiao Li, and Guanyi Chen.
  2018{\natexlab{b}}.
\newblock A dual-attention hierarchical recurrent neural network for dialogue
  act classification.
\newblock \emph{CoRR}.

\bibitem[{Li et~al.(2018{\natexlab{c}})Li, Lin, Collinson, Li, and
  Chen}]{crf_li}
Ruizhe Li, Chenghua Lin, Matthew Collinson, Xiao Li, and Guanyi Chen.
  2018{\natexlab{c}}.
\newblock A dual-attention hierarchical recurrent neural network for dialogue
  act classification.
\newblock \emph{arXiv preprint arXiv:1810.09154}.

\bibitem[{Li et~al.(2017)Li, Su, Shen, Li, Cao, and Niu}]{dataset_dailydialog}
Yanran Li, Hui Su, Xiaoyu Shen, Wenjie Li, Ziqiang Cao, and Shuzi Niu. 2017.
\newblock \href {http://arxiv.org/abs/1710.03957} {Dailydialog: A manually
  labelled multi-turn dialogue dataset}.

\bibitem[{Lin et~al.(2017)Lin, Feng, Santos, Yu, Xiang, Zhou, and
  Bengio}]{self_attention}
Zhouhan Lin, Minwei Feng, Cicero Nogueira~dos Santos, Mo~Yu, Bing Xiang, Bowen
  Zhou, and Yoshua Bengio. 2017.
\newblock A structured self-attentive sentence embedding.
\newblock \emph{arXiv preprint arXiv:1703.03130}.

\bibitem[{Lison and Tiedemann(2016)}]{opensub}
Pierre Lison and J{\"o}rg Tiedemann. 2016.
\newblock Opensubtitles2016: Extracting large parallel corpora from movie and
  tv subtitles.

\bibitem[{Lison et~al.(2019)Lison, Tiedemann, Kouylekov et~al.}]{open}
Pierre Lison, J{\"o}rg Tiedemann, Milen Kouylekov, et~al. 2019.
\newblock Open subtitles 2018: Statistical rescoring of sentence alignments in
  large, noisy parallel corpora.
\newblock In \emph{LREC 2018, Eleventh International Conference on Language
  Resources and Evaluation}. European Language Resources Association (ELRA).

\bibitem[{Liu et~al.(2019)Liu, Ott, Goyal, Du, Joshi, Chen, Levy, Lewis,
  Zettlemoyer, and Stoyanov}]{roberta}
Yinhan Liu, Myle Ott, Naman Goyal, Jingfei Du, Mandar Joshi, Danqi Chen, Omer
  Levy, Mike Lewis, Luke Zettlemoyer, and Veselin Stoyanov. 2019.
\newblock Roberta: A robustly optimized bert pretraining approach.
\newblock \emph{arXiv preprint arXiv:1907.11692}.

\bibitem[{Loshchilov and Hutter(2017)}]{adamW}
Ilya Loshchilov and Frank Hutter. 2017.
\newblock Decoupled weight decay regularization.
\newblock \emph{arXiv preprint arXiv:1711.05101}.

\bibitem[{Lowe et~al.(2015)Lowe, Pow, Serban, and Pineau}]{ubuntu}
Ryan Lowe, Nissan Pow, Iulian Serban, and Joelle Pineau. 2015.
\newblock \href {http://arxiv.org/abs/1506.08909} {The ubuntu dialogue corpus:
  {A} large dataset for research in unstructured multi-turn dialogue systems}.
\newblock \emph{CoRR}, abs/1506.08909.

\bibitem[{Mckeown et~al.(2013)Mckeown, Valstar, Cowie, Pantic, and
  Schroder}]{dataset_semaine}
Gary Mckeown, Michel Valstar, Roddy Cowie, Maja Pantic, and M.~Schroder. 2013.
\newblock \href {https://doi.org/10.1109/T-AFFC.2011.20} {The semaine database:
  Annotated multimodal records of emotionally colored conversations between a
  person and a limited agent}.
\newblock \emph{Affective Computing, IEEE Transactions on}, 3:5--17.

\bibitem[{Mehri et~al.(2019)Mehri, Razumovsakaia, Zhao, and
  Eskenazi}]{mehri2019pretraining}
Shikib Mehri, Evgeniia Razumovsakaia, Tiancheng Zhao, and Maxine Eskenazi.
  2019.
\newblock Pretraining methods for dialog context representation learning.
\newblock \emph{arXiv preprint arXiv:1906.00414}.

\bibitem[{Mikolov et~al.(2013)Mikolov, Sutskever, Chen, Corrado, and
  Dean}]{mikolov2013distributed}
Tomas Mikolov, Ilya Sutskever, Kai Chen, Greg~S Corrado, and Jeff Dean. 2013.
\newblock Distributed representations of words and phrases and their
  compositionality.
\newblock In \emph{Advances in neural information processing systems}, pages
  3111--3119.

\bibitem[{Passonneau and Sachar.(2014)}]{dataset_loquihuman}
R.~Passonneau and E.~Sachar. 2014.
\newblock Loqui human-human dialogue corpus (transcriptions and annotations).

\bibitem[{Pennington et~al.(2014)Pennington, Socher, and
  Manning}]{pennington2014glove}
Jeffrey Pennington, Richard Socher, and Christopher~D Manning. 2014.
\newblock Glove: Global vectors for word representation.
\newblock In \emph{Proceedings of the 2014 conference on empirical methods in
  natural language processing (EMNLP)}, pages 1532--1543.

\bibitem[{Peters et~al.(2018)Peters, Neumann, Iyyer, Gardner, Clark, Lee, and
  Zettlemoyer}]{peters2018deep}
Matthew~E Peters, Mark Neumann, Mohit Iyyer, Matt Gardner, Christopher Clark,
  Kenton Lee, and Luke Zettlemoyer. 2018.
\newblock Deep contextualized word representations.
\newblock \emph{arXiv preprint arXiv:1802.05365}.

\bibitem[{Poria et~al.(2018{\natexlab{a}})Poria, Hazarika, Majumder, Naik,
  Cambria, and Mihalcea}]{dataset_meld}
Soujanya Poria, Devamanyu Hazarika, Navonil Majumder, Gautam Naik, Erik
  Cambria, and Rada Mihalcea. 2018{\natexlab{a}}.
\newblock \href {http://arxiv.org/abs/1810.02508} {Meld: A multimodal
  multi-party dataset for emotion recognition in conversations}.

\bibitem[{Poria et~al.(2018{\natexlab{b}})Poria, Hazarika, Majumder, Naik,
  Cambria, and Mihalcea}]{weighted_no}
Soujanya Poria, Devamanyu Hazarika, Navonil Majumder, Gautam Naik, Erik
  Cambria, and Rada Mihalcea. 2018{\natexlab{b}}.
\newblock Meld: A multimodal multi-party dataset for emotion recognition in
  conversations.
\newblock \emph{arXiv preprint arXiv:1810.02508}.

\bibitem[{Ruder(2017)}]{multi_task_2}
Sebastian Ruder. 2017.
\newblock An overview of multi-task learning in deep neural networks.
\newblock \emph{arXiv preprint arXiv:1706.05098}.

\bibitem[{Sanh et~al.(2019)Sanh, Wolf, and Ruder}]{hierarchy_loss}
Victor Sanh, Thomas Wolf, and Sebastian Ruder. 2019.
\newblock A hierarchical multi-task approach for learning embeddings from
  semantic tasks.
\newblock In \emph{Proceedings of the AAAI Conference on Artificial
  Intelligence}, volume~33, pages 6949--6956.

\bibitem[{Sener and Koltun(2018)}]{multi_loss_opt}
Ozan Sener and Vladlen Koltun. 2018.
\newblock Multi-task learning as multi-objective optimization.
\newblock In \emph{Advances in Neural Information Processing Systems}, pages
  527--538.

\bibitem[{Serban et~al.(2015)Serban, Sordoni, Bengio, Courville, and
  Pineau}]{hred}
Iulian~Vlad Serban, Alessandro Sordoni, Yoshua Bengio, Aaron~C. Courville, and
  Joelle Pineau. 2015.
\newblock \href {http://arxiv.org/abs/1507.04808} {Hierarchical neural network
  generative models for movie dialogues}.
\newblock \emph{CoRR}, abs/1507.04808.

\bibitem[{Shriberg et~al.(2004)Shriberg, Dhillon, Bhagat, Ang, and
  Carvey}]{dataset_mrda}
Elizabeth Shriberg, Raj Dhillon, Sonali Bhagat, Jeremy Ang, and Hannah Carvey.
  2004.
\newblock \href {https://www.aclweb.org/anthology/W04-2319} {The {ICSI} meeting
  recorder dialog act ({MRDA}) corpus}.
\newblock In \emph{Proceedings of the 5th {SIG}dial Workshop on Discourse and
  Dialogue at {HLT}-{NAACL} 2004}, pages 97--100, Cambridge, Massachusetts,
  USA. Association for Computational Linguistics.

\bibitem[{Shriberg(1999)}]{fillers}
Elizabeth~E Shriberg. 1999.
\newblock Phonetic consequences of speech disfluency.
\newblock Technical report, SRI INTERNATIONAL MENLO PARK CA.

\bibitem[{Srivastava et~al.(2014)Srivastava, Hinton, Krizhevsky, Sutskever, and
  Salakhutdinov}]{dropout}
Nitish Srivastava, Geoffrey Hinton, Alex Krizhevsky, Ilya Sutskever, and Ruslan
  Salakhutdinov. 2014.
\newblock Dropout: a simple way to prevent neural networks from overfitting.
\newblock \emph{The journal of machine learning research}, 15(1):1929--1958.

\bibitem[{Staerman et~al.(2020{\natexlab{a}})Staerman, Laforgue, Mozharovskyi,
  and d'Alch{\'e} Buc}]{robust_staerman}
Guillaume Staerman, Pierre Laforgue, Pavlo Mozharovskyi, and Florence
  d'Alch{\'e} Buc. 2020{\natexlab{a}}.
\newblock When ot meets mom: Robust estimation of wasserstein distance.
\newblock \emph{arXiv preprint arXiv:2006.10325}.

\bibitem[{Staerman et~al.(2020{\natexlab{b}})Staerman, Mozharovskyi, Cl{\'e}men
  et~al.}]{anomaly_2}
Guillaume Staerman, Pavlo Mozharovskyi, St{\'e}phan Cl{\'e}men, et~al.
  2020{\natexlab{b}}.
\newblock The area of the convex hull of sampled curves: a robust functional
  statistical depth measure.
\newblock In \emph{International Conference on Artificial Intelligence and
  Statistics}, pages 570--579.

\bibitem[{Staerman et~al.(2019)Staerman, Mozharovskyi, Cl{\'e}men{\c{c}}on, and
  d'Alch{\'e} Buc}]{anomaly_1}
Guillaume Staerman, Pavlo Mozharovskyi, Stephan Cl{\'e}men{\c{c}}on, and
  Florence d'Alch{\'e} Buc. 2019.
\newblock Functional isolation forest.
\newblock \emph{arXiv preprint arXiv:1904.04573}.

\bibitem[{Stolcke et~al.(2000)Stolcke, Ries, Coccaro, Shriberg, Bates,
  Jurafsky, Taylor, Martin, Ess-Dykema, and Meteer}]{hmm_dialog}
Andreas Stolcke, Klaus Ries, Noah Coccaro, Elizabeth Shriberg, Rebecca Bates,
  Daniel Jurafsky, Paul Taylor, Rachel Martin, Carol~Van Ess-Dykema, and Marie
  Meteer. 2000.
\newblock Dialogue act modeling for automatic tagging and recognition of
  conversational speech.
\newblock \emph{Computational linguistics}, 26(3):339--373.

\bibitem[{Stolcke and Shriberg(1996)}]{disfluency}
Andreas Stolcke and Elizabeth Shriberg. 1996.
\newblock Statistical language modeling for speech disfluencies.
\newblock In \emph{1996 IEEE International Conference on Acoustics, Speech, and
  Signal Processing Conference Proceedings}, volume~1, pages 405--408. IEEE.

\bibitem[{Su{\'a}rez et~al.(2019)Su{\'a}rez, Sagot, and Romary}]{oscar}
Pedro Javier~Ortiz Su{\'a}rez, Beno{\^\i}t Sagot, and Laurent Romary. 2019.
\newblock Asynchronous pipeline for processing huge corpora on medium to low
  resource infrastructures.
\newblock \emph{Challenges in the Management of Large Corpora (CMLC-7) 2019},
  page~9.

\bibitem[{Surendran and Levow(2006)}]{svm_dialog}
Dinoj Surendran and Gina-Anne Levow. 2006.
\newblock Dialog act tagging with support vector machines and hidden markov
  models.
\newblock In \emph{Ninth International Conference on Spoken Language
  Processing}.

\bibitem[{Thompson et~al.(1993)Thompson, Anderson, Bard, Doherty-Sneddon,
  Newlands, and Sotillo}]{dataset_maptask}
Henry Thompson, Anne Anderson, Ellen Bard, Gwyneth Doherty-Sneddon, Alison
  Newlands, and Cathy Sotillo. 1993.
\newblock \href {https://doi.org/10.3115/1075671.1075677} {The hcrc map task
  corpus: natural dialogue for speech recognition}.

\bibitem[{Thornbury and Slade(2006)}]{context}
Scott Thornbury and Diana Slade. 2006.
\newblock \emph{Conversation: From description to pedagogy}.
\newblock Cambridge University Press.

\bibitem[{Tran et~al.(2017)Tran, Haffari, and Zukerman}]{tran2017generative}
Quan~Hung Tran, Gholamreza Haffari, and Ingrid Zukerman. 2017.
\newblock A generative attentional neural network model for dialogue act
  classification.
\newblock In \emph{Proceedings of the 55th Annual Meeting of the Association
  for Computational Linguistics (Volume 2: Short Papers)}, pages 524--529.

\bibitem[{Tran et~al.(2019)Tran, Yuan, Liu, and Ostendorf}]{tran2019role}
Trang Tran, Jiahong Yuan, Yang Liu, and Mari Ostendorf. 2019.
\newblock On the role of style in parsing speech with neural models.
\newblock \emph{Proc. Interspeech 2019}, pages 4190--4194.

\bibitem[{Vaswani et~al.(2017)Vaswani, Shazeer, Parmar, Uszkoreit, Jones,
  Gomez, Kaiser, and Polosukhin}]{attention_is}
Ashish Vaswani, Noam Shazeer, Niki Parmar, Jakob Uszkoreit, Llion Jones,
  Aidan~N Gomez, {\L}ukasz Kaiser, and Illia Polosukhin. 2017.
\newblock Attention is all you need.
\newblock In \emph{Advances in neural information processing systems}, pages
  5998--6008.

\bibitem[{Wang et~al.(2018)Wang, Singh, Michael, Hill, Levy, and Bowman}]{glue}
Alex Wang, Amanpreet Singh, Julian Michael, Felix Hill, Omer Levy, and Samuel~R
  Bowman. 2018.
\newblock Glue: A multi-task benchmark and analysis platform for natural
  language understanding.
\newblock \emph{arXiv preprint arXiv:1804.07461}.

\bibitem[{Witon et~al.(2018)Witon, Colombo, Modi, and Kapadia}]{classif}
Wojciech Witon, Pierre Colombo, Ashutosh Modi, and Mubbasir Kapadia. 2018.
\newblock Disney at iest 2018: Predicting emotions using an ensemble.
\newblock In \emph{Proceedings of the 9th Workshop on Computational Approaches
  to Subjectivity, Sentiment and Social Media Analysis}, pages 248--253.

\bibitem[{Wolf et~al.(2019)Wolf, Debut, Sanh, Chaumond, Delangue, Moi, Cistac,
  Rault, Louf, Funtowicz, and Brew}]{Wolf2019HuggingFacesTS}
Thomas Wolf, Lysandre Debut, Victor Sanh, Julien Chaumond, Clement Delangue,
  Anthony Moi, Pierric Cistac, Tim Rault, R'emi Louf, Morgan Funtowicz, and
  Jamie Brew. 2019.
\newblock Huggingface's transformers: State-of-the-art natural language
  processing.
\newblock \emph{ArXiv}, abs/1910.03771.

\bibitem[{Wolf et~al.(2020)Wolf, Lhoest, von Platen, Jernite, Drame, Plu,
  Chaumond, Delangue, Ma, Thakur, Patil, Davison, Scao, Sanh, Xu, Patry,
  McMillan-Major, Brandeis, Gugger, Lagunas, Debut, Funtowicz, Moi, Rush,
  Schmidd, Cistac, Muštar, Boudier, and Tordjmann}]{2020HuggingFace-datasets}
Thomas Wolf, Quentin Lhoest, Patrick von Platen, Yacine Jernite, Mariama Drame,
  Julien Plu, Julien Chaumond, Clement Delangue, Clara Ma, Abhishek Thakur,
  Suraj Patil, Joe Davison, Teven~Le Scao, Victor Sanh, Canwen Xu, Nicolas
  Patry, Angie McMillan-Major, Simon Brandeis, Sylvain Gugger, François
  Lagunas, Lysandre Debut, Morgan Funtowicz, Anthony Moi, Sasha Rush, Philipp
  Schmidd, Pierric Cistac, Victor Muštar, Jeff Boudier, and Anna Tordjmann.
  2020.
\newblock Datasets.
\newblock \emph{GitHub. Note: https://github.com/huggingface/datasets}, 1.

\bibitem[{Wu et~al.(2016)Wu, Schuster, Chen, Le, Norouzi, Macherey, Krikun,
  Cao, Gao, Macherey et~al.}]{wordpiece}
Yonghui Wu, Mike Schuster, Zhifeng Chen, Quoc~V Le, Mohammad Norouzi, Wolfgang
  Macherey, Maxim Krikun, Yuan Cao, Qin Gao, Klaus Macherey, et~al. 2016.
\newblock Google's neural machine translation system: Bridging the gap between
  human and machine translation.
\newblock \emph{arXiv preprint arXiv:1609.08144}.

\bibitem[{Yang et~al.(2019)Yang, Dai, Yang, Carbonell, Salakhutdinov, and
  Le}]{xlnet}
Zhilin Yang, Zihang Dai, Yiming Yang, Jaime Carbonell, Russ~R Salakhutdinov,
  and Quoc~V Le. 2019.
\newblock Xlnet: Generalized autoregressive pretraining for language
  understanding.
\newblock In \emph{Advances in neural information processing systems}, pages
  5754--5764.

\bibitem[{Yi et~al.(2019)Yi, Goel, Khatri, Cervone, Chung, Hedayatnia,
  Venkatesh, Gabriel, and Hakkani-Tur}]{generic}
Sanghyun Yi, Rahul Goel, Chandra Khatri, Alessandra Cervone, Tagyoung Chung,
  Behnam Hedayatnia, Anu Venkatesh, Raefer Gabriel, and Dilek Hakkani-Tur.
  2019.
\newblock Towards coherent and engaging spoken dialog response generation using
  automatic conversation evaluators.
\newblock \emph{arXiv preprint arXiv:1904.13015}.

\bibitem[{Zhang et~al.(2019{\natexlab{a}})Zhang, Wei, and
  Zhou}]{zhang2019hibert}
Xingxing Zhang, Furu Wei, and Ming Zhou. 2019{\natexlab{a}}.
\newblock Hibert: Document level pre-training of hierarchical bidirectional
  transformers for document summarization.
\newblock \emph{arXiv preprint arXiv:1905.06566}.

\bibitem[{Zhang et~al.(2019{\natexlab{b}})Zhang, Li, Song, Zhang, and
  Wang}]{accuracy_fscore}
Yazhou Zhang, Qiuchi Li, Dawei Song, Peng Zhang, and Panpan Wang.
  2019{\natexlab{b}}.
\newblock Quantum-inspired interactive networks for conversational sentiment
  analysis.

\bibitem[{Zhu et~al.(2015)Zhu, Kiros, Zemel, Salakhutdinov, Urtasun, Torralba,
  and Fidler}]{book}
Yukun Zhu, Ryan Kiros, Rich Zemel, Ruslan Salakhutdinov, Raquel Urtasun,
  Antonio Torralba, and Sanja Fidler. 2015.
\newblock Aligning books and movies: Towards story-like visual explanations by
  watching movies and reading books.
\newblock In \emph{Proceedings of the IEEE international conference on computer
  vision}, pages 19--27.

\end{thebibliography}
\bibliographystyle{acl_natbib}
\appendix
\clearpage

\section{Additional Details on data composing \texttt{SILICONE}}\label{supp:copora}
\begin{figure*}
    \centering
    \begin{subfigure}[b]{0.45\textwidth}   
        \centering 
        \includegraphics[width=\textwidth]{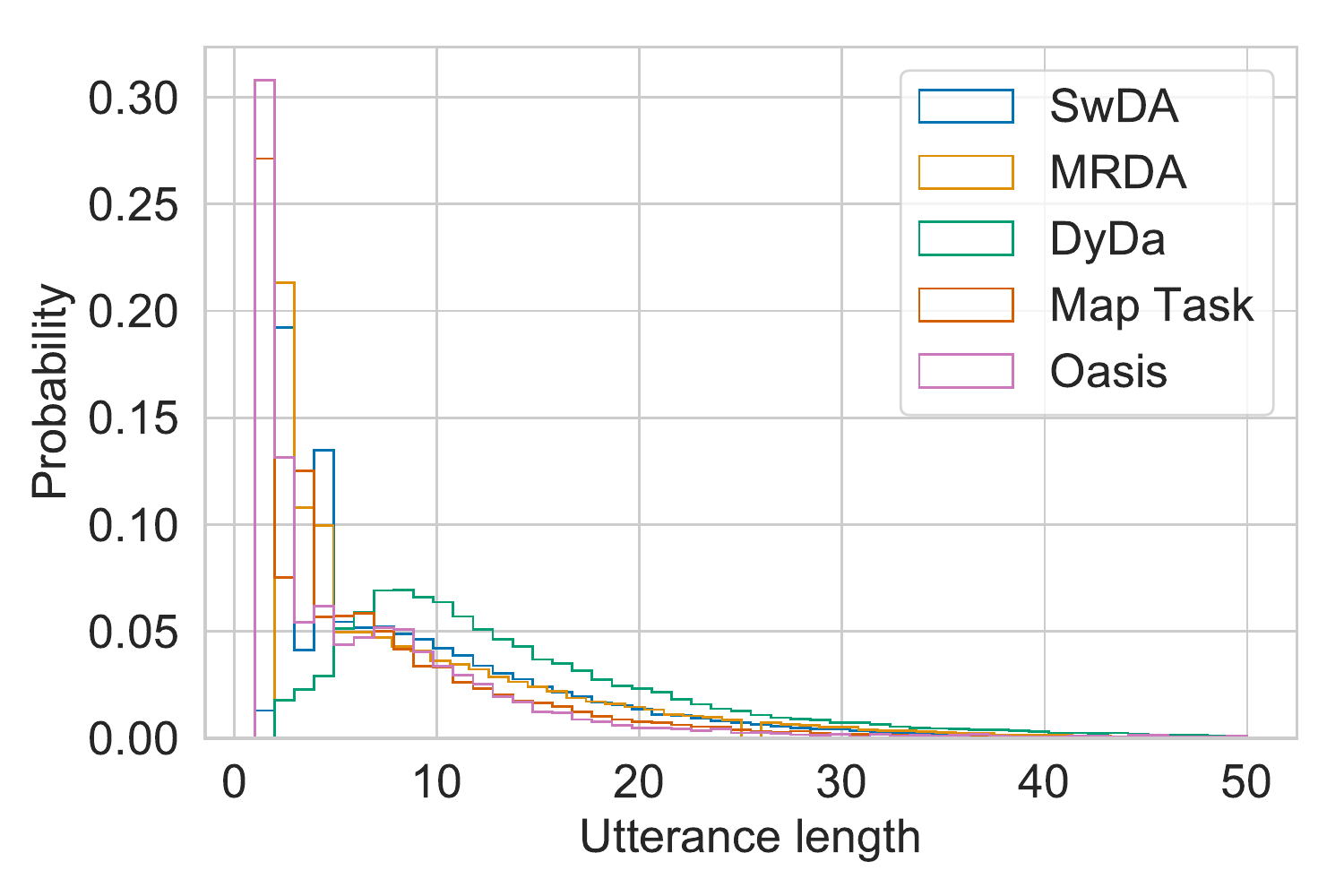}
        \caption{\texttt{SILICONE} \texttt{DA}}       
        \label{fig:mean and std of net34}
    \end{subfigure}
    \quad
    \begin{subfigure}[b]{0.45\textwidth}   
        \centering 
        \includegraphics[width=\textwidth]{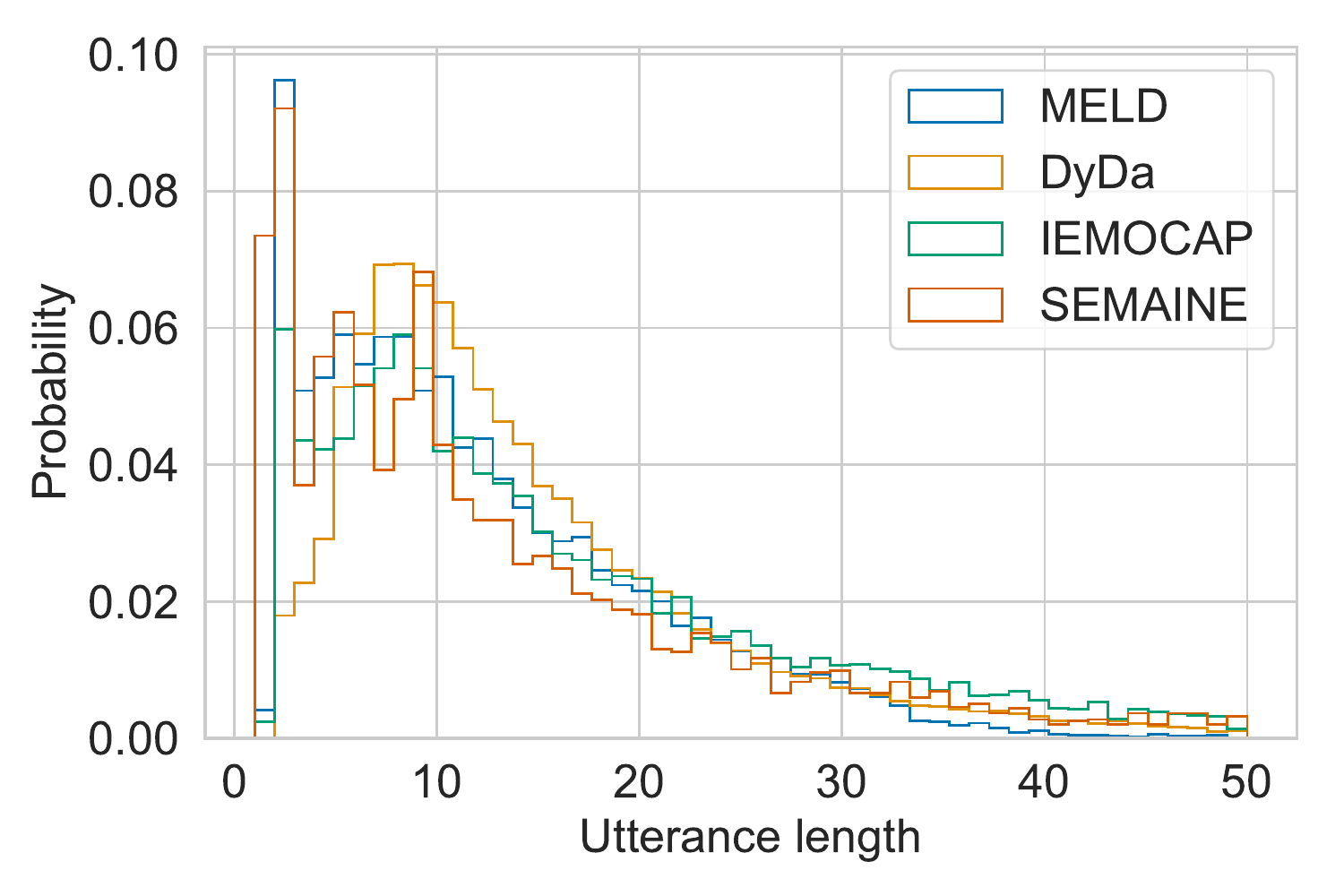}
        \caption{\texttt{SILICONE} \texttt{S/E}}       
        \label{fig:mean and std of net44}
    \end{subfigure}
    \vskip\baselineskip
    \caption{Histograms showing the utterance length for each dataset of \texttt{SILICONE}.} 
    \label{fig:lght_utt}
\end{figure*}

In this section, we illustrate the diversity of the dataset composing \texttt{SILICONE}. In~\autoref{fig:lght_utt}, we plot two histograms representing the different utterance lengths for \texttt{DA} and \texttt{E/S}. As expected, for spoken dialog, lengths are shorter than for written benchmarks (e.g., GLUE).

\section{Additional Details for Models}
\label{sec:sup_model}
In this section we report model hyper-parameters and as well as additional descriptions of our baselines. For all models we use a tokenizer based on WordPiece~\cite{wordpiece}.\\
We also provide a concrete example of corrupted context for the \texttt{MLM} Loss.

\subsection{Hierarchical pre-training}
We report in~\autoref{tab:archi_hyper} the main hyper-parameters used fo our model pre-training. We used GELU~\cite{gelu} activations and the dropout rate \cite{dropout} is set to $0.1$.
\begin{table}[]
    \centering
    \begin{tabular}{c|cc}\hline
     & \texttt{TINY}& \texttt{SMALL}  \\\hline
     Nbs of heads   & 1& 6  \\
      $N_d$  & 2&4 \\
      $N_u$ &2 & 4\\
      $T$ & 50 & 50\\
      $C$ & 5 & 5\\
      $\mathcal{T}_d$ nbs of heads  &6 &6 \\
      Inner dimension  &768 &768 \\
  Model Dimension   &768 &768 \\
Vocab length  &32000 &32000 \\
  $\mathcal{T}_d$: Emb. size& 768 &768 \\
  $d_k$:&64 & 64\\
  $d_v$: &64 & 64
    \end{tabular}
    \caption{Architecture hyperparameters used for the hierarchical pre-training.}
    \label{tab:archi_hyper}
\end{table}
\subsection{\texttt{MLM} Loss example}
\begin{figure*}
    \centering
            \begin{subfigure}[b]{0.45\textwidth}   
        \centering 
        \includegraphics[width=\textwidth]{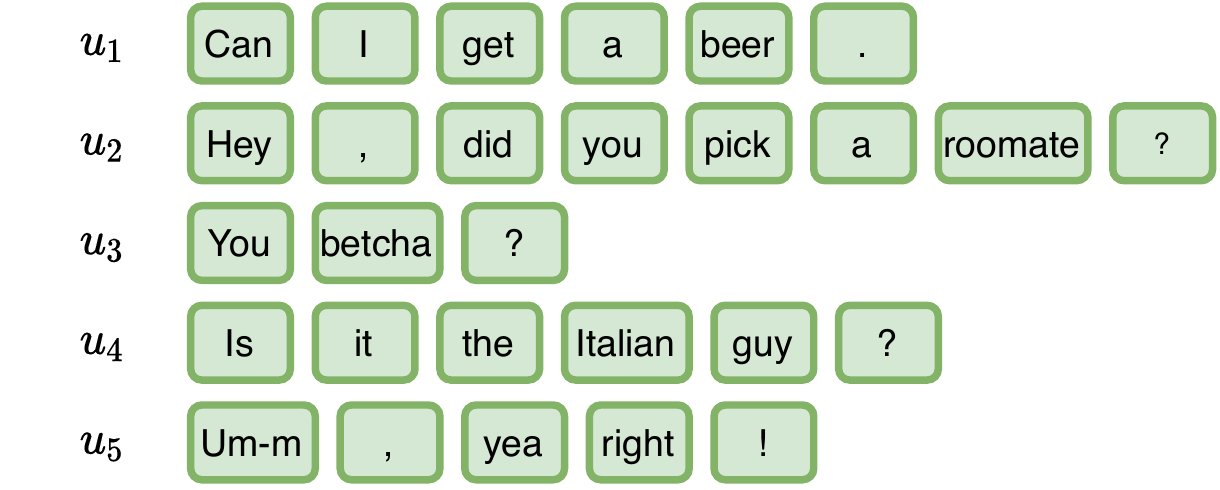}
        \caption{Initial context composed by 5 utterances.}       
        \label{fig:corrup_exp_initial_ctx}
    \end{subfigure}
    \vskip\baselineskip
    \begin{subfigure}[b]{0.45\textwidth}   
        \centering 
        \includegraphics[width=\textwidth]{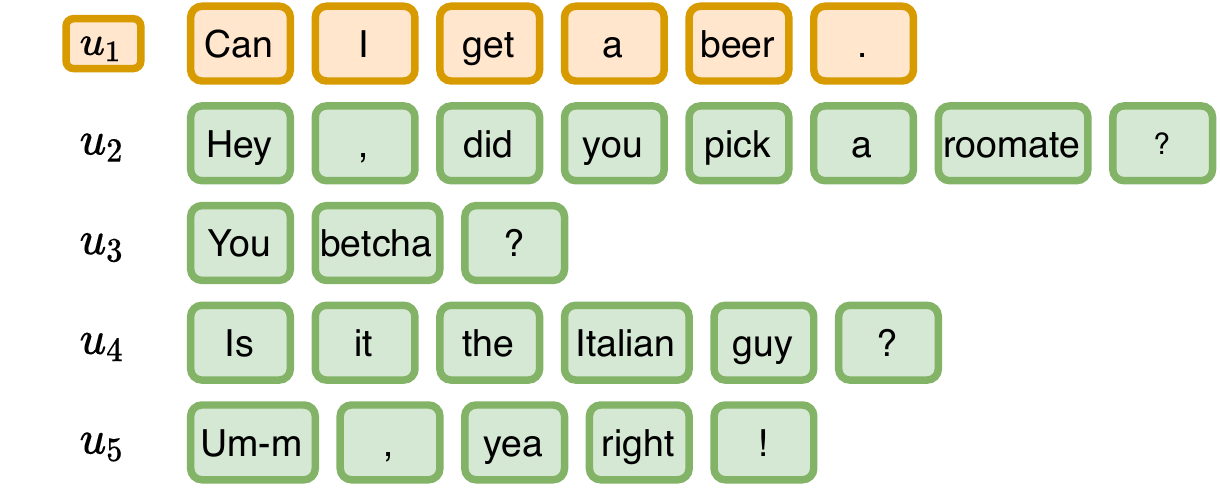}
        \caption{$u_1$ is chosen to be masked.}       
        \label{fig:corrup_exp_pick_utt_1}
    \end{subfigure}
    \quad     
    \begin{subfigure}[b]{0.45\textwidth}  
        \centering 
        \includegraphics[width=\textwidth]{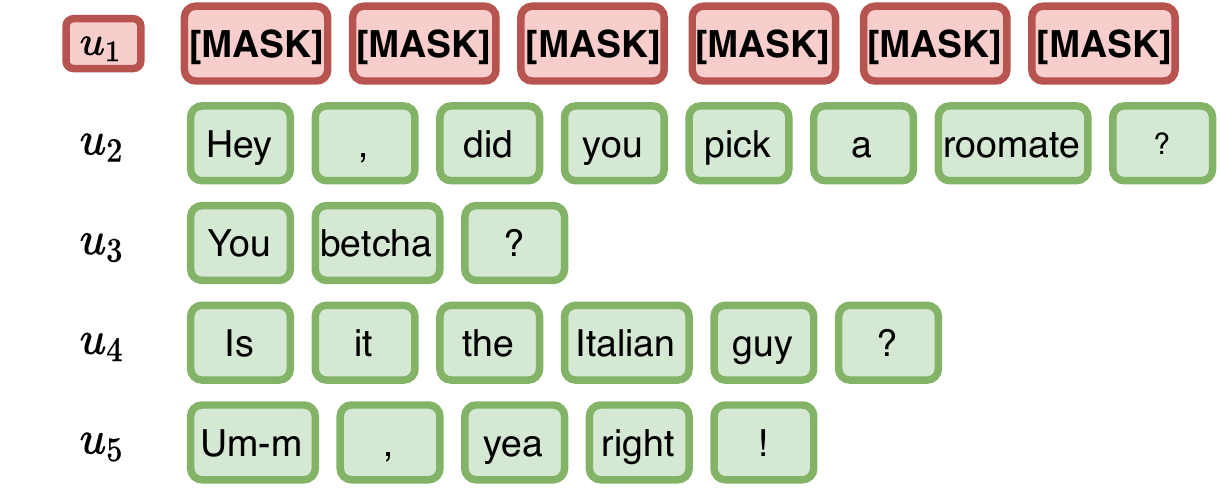}
        \caption{Corrupted context with utterance $u_1$ masked.}    
        \label{fig:corrup_exp_mask_utt_1}
    \end{subfigure}
    \vskip\baselineskip
    \begin{subfigure}[b]{0.45\textwidth}   
        \centering 
        \includegraphics[width=\textwidth]{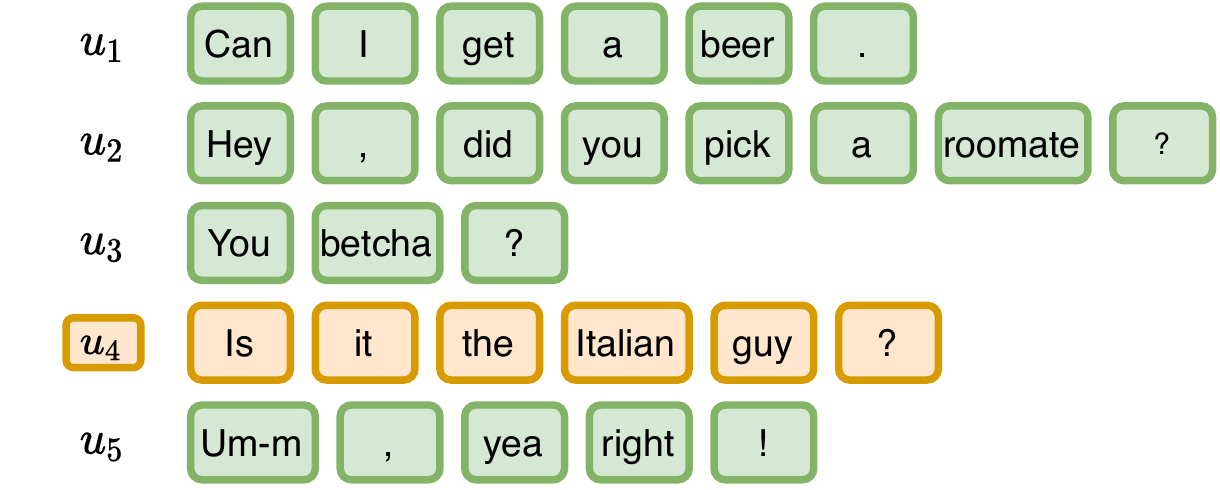}
        \caption{$u_4$ is chosen to be masked.}       
        \label{fig:corrup_exp_pick_utt_2}
    \end{subfigure}
     \quad       
    \begin{subfigure}[b]{0.45\textwidth}  
        \centering 
        \includegraphics[width=\textwidth]{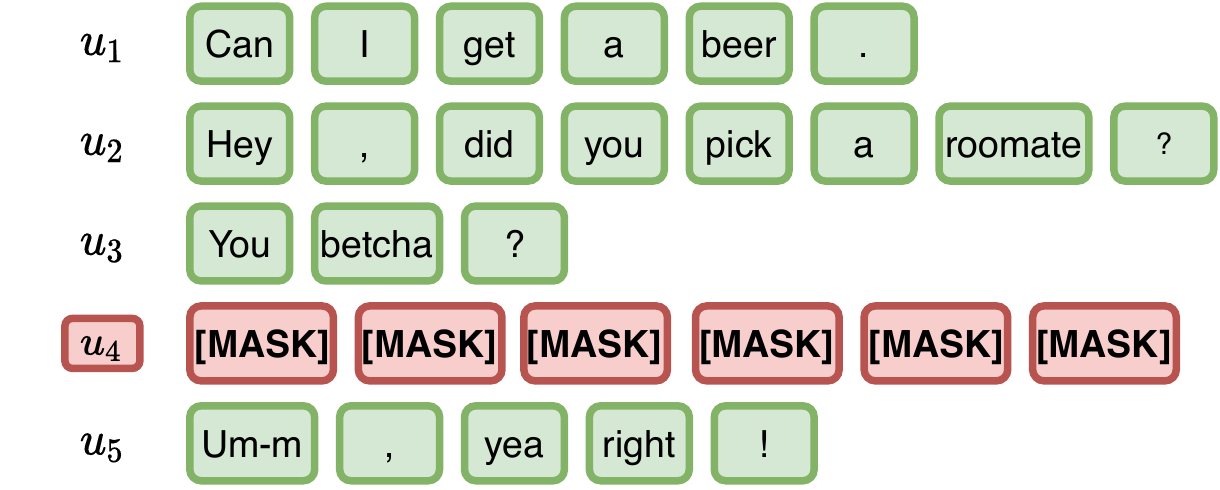}
        \caption{Corrupted context with utterance $u_4$ masked.}    
        \label{fig:corrup_exp_mask_utt_2}
    \end{subfigure}
    \caption{This figure shows an example of corrupted context. Here $p_C$ is randmoly set to 2 meaning that two utterances will be corrupted. $u_1$ and $u_4$ are randomly picked in \ref{fig:corrup_exp_pick_utt_1}, \ref{fig:corrup_exp_pick_utt_2} and then masked in  \ref{fig:corrup_exp_mask_utt_1}, \ref{fig:corrup_exp_mask_utt_2}.
    }
    \label{fig:corrup_exp}
\end{figure*}

In this section we propose a visual illustration of the corrupted context \autoref{fig:corrup_exp} by the \texttt{MLM} Loss.

\subsection{Experimental Hyper-parameters for \texttt{SILICONE}}
For all models, we use a batch size of $64$ and automatically select the best model on the validation set according to its loss.
We do not perform exhaustive grid search either on the learning rate (that is set to $10^{-4}$), nor on other hyper-parameters to perform a fair comparison between all the models. We use ADAMW~\cite{adam,adamW} with a linear scheduler on the learning rate and the number of warm-up steps is set to $100$.

\subsection{Additional Details on Baselines}
\begin{figure*}
    \centering
            \begin{subfigure}[b]{0.475\textwidth}   
        \centering 
        \includegraphics[width=\textwidth]{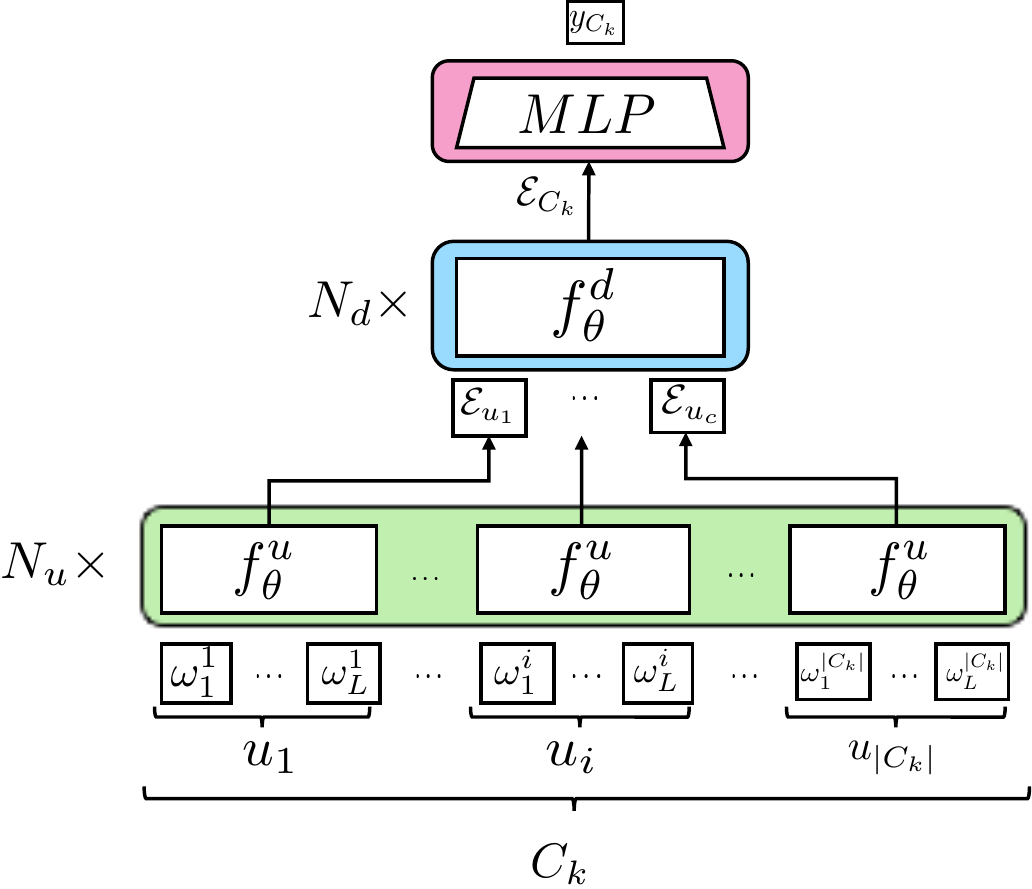}
        \caption{Hierarchical encoder with MLP decoder performing single label prediction.}       
        \label{fig:h_mlp}
    \end{subfigure}
    \quad
    \begin{subfigure}[b]{0.41\textwidth}   
        \centering 
        \includegraphics[width=\textwidth]{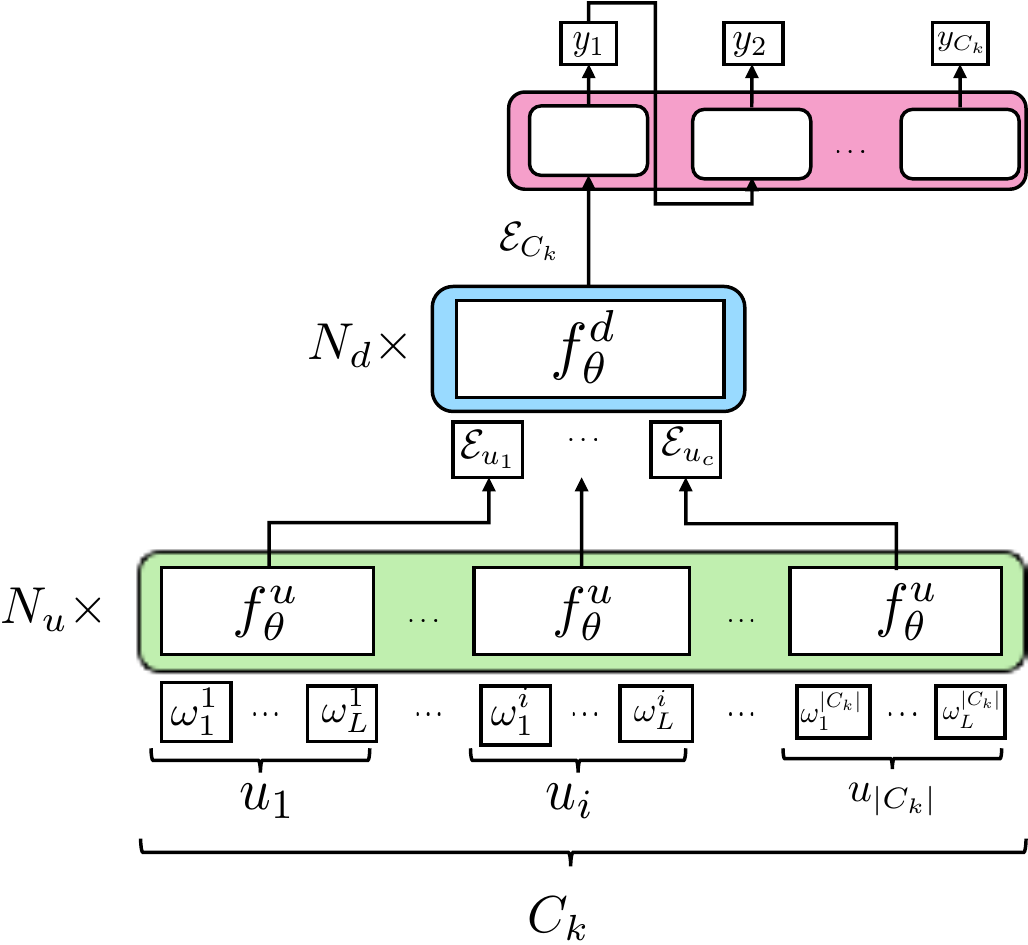}
        \caption{Hierarchical encoder with sequential decoder (either GRU or CRF).}       
        \label{fig:h_seq}
    \end{subfigure}
            \vskip\baselineskip
            
    \begin{subfigure}[b]{0.475\textwidth}  
        \centering 
        \includegraphics[width=\textwidth]{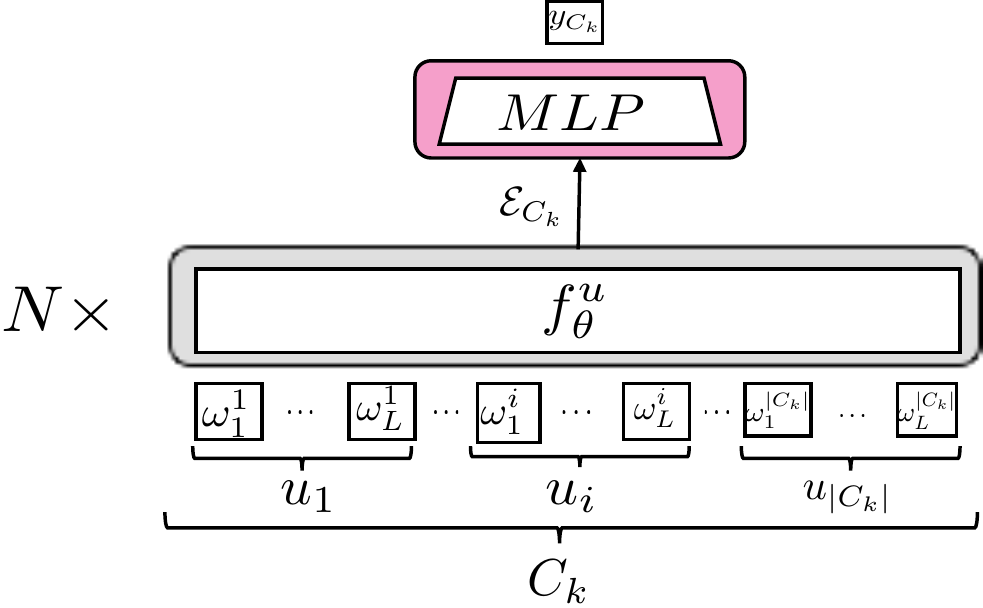}
        \caption{BERT encoder with MLP decoder performing single label prediction.}    
        \label{fig:b_mlp}
    \end{subfigure}
    \hfill
    \begin{subfigure}[b]{0.475\textwidth}
        \centering
        \includegraphics[width=\textwidth]{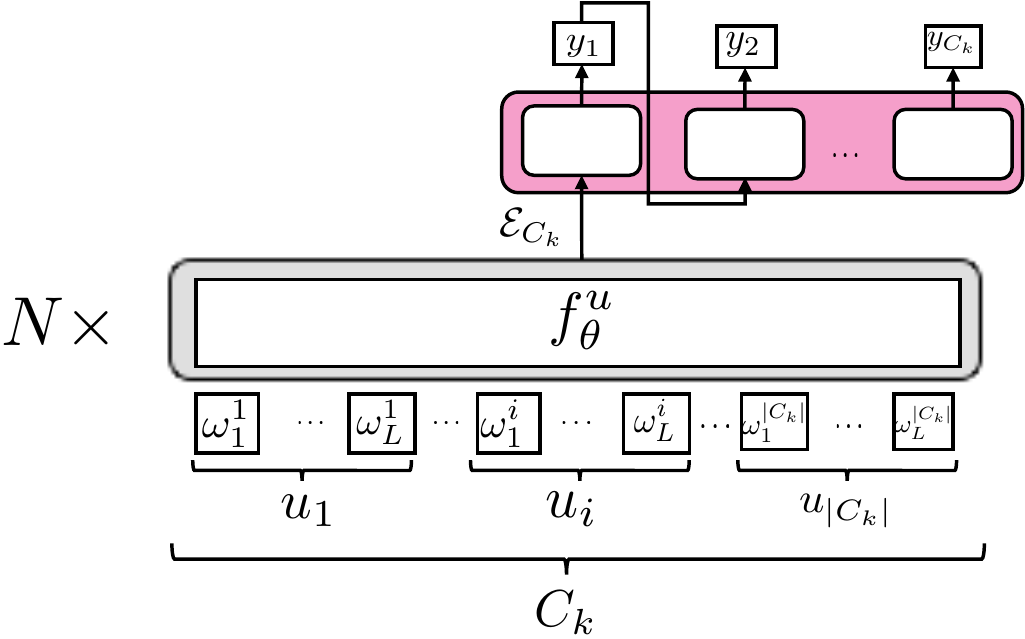}
        \caption{BERT encoder with sequential decoder (either GRU or CRF)}    
        \label{fig:b_seq}
    \end{subfigure}

    \caption{Schema of the different models evaluated on \texttt{SILICONE}. In this figure $f^u_\theta$, $f^d_\theta$ and the sequence label decoder ($g_\theta^{dec}$) are respectively colored in green, blue and red for the hierarchical encoder (see \autoref{fig:h_mlp} and \autoref{fig:b_seq}). For BERT there is no hierarchy and embedding is performed through $f^u_\theta$ colored in grey (see \autoref{fig:b_mlp}, \autoref{fig:b_seq})} 
    \label{fig:baselines_schema_powe}
\end{figure*}

A representation for all the baselines can be found in~\autoref{fig:baselines_schema_powe}. For all models, both hidden dimension and embedding dimension is set to $768$ to ensure fair comparison with the proposed model. The MLP used for decoding contains 3 layers of sizes $(768,348,192)$. We use RELU~\cite{relu} to introduce non linearity inside our architecture.

\section{Additional Experimental Results}
\label{supp:results}
In this section we report the detailed results on \texttt{SILICONE}, including the ones presented in~\autoref{tab:results}. We report results on two new experiments: importance of pre-training time for both a TINY and SMALL model, we report the convergence time of a TINY model and finally we extend~\autoref{sssec:multi_sup} by reporting results on \texttt{IEMO}.

\subsection{Detailed Results on \texttt{SILICONE}}
We show in~\autoref{tab:full_results} the results on the \texttt{SILICONE} benchmark for all the models mentioned in the paper.

\begin{table*}
\begin{center}
 \resizebox{\textwidth}{!}{\begin{tabular}{c || c || c c c  c c | c c c c c } 
 \hline
 & \textbf{Avg} & \texttt{SwDA}& \texttt{MRDA} &  $\texttt{DyDA}_{\texttt{DA}}$ & \texttt{MT} & \texttt{Oasis} & $\texttt{DyDA}_\texttt{e}$  & $\texttt{MELD}_\texttt{s}$  & $\texttt{MELD}_\texttt{e}$ & \texttt{IEMO} &  \texttt{SEM}  \\ \hline
      BERT-4layers (+MLP)& 69.45 & 77.8 & 90.7 & 79.0  &88.4 & 66.8 & 90.3 & 49.3  &50.4 & 43.0 &58.8\\
    BERT (+MLP)& 72.79 & 79.2  &90.7 & 82.6 & 88.2   &66.9 & 91.9 & 59.3  &61.4 & 45.0 &62.7\\
    BERT (+GRU)& 69.84 & 78.2  &90.4 & 80.8 & 88.7   &63.7 & 90 & 50.4  &48.9 & 45.0 &62.3\\
        BERT (+CRF)& 72.8 & 79.0  &90.8 & 88.3   &67.2  &   81.9& 91.5 & 59.4  &61.0 & 44.2 &61.5\\ \hline
     $\mathcal{H}\mathcal{R}$ (+MLP)& 69.77 & 77,5  &90,9 & 80,1 & 82,8 & 64,3 & 91.5  &59,3 & 59.9 & 40.3& 51.1\\
     $\mathcal{H}\mathcal{R}$ (+GRU)&67.54 & 78.2 & 90.9 & 79,9 & 84,4   &63,5 & 91.5& 50,7 & 50.4 & 35.2 & 50.7  \\
     $\mathcal{H}\mathcal{R}$ (+CRF)& 70.5 & 77.8  &91,3 & 79,7 & 87,5   & 65,3  & 91,1  &62,1 & 57,4  & 42.1 &50.7\\ \hline
     $\mathcal{H}\mathcal{T}(\theta_{MLM}^{u,d})$ (TINY)  &73.3& 79.3  &92.0 & 80.1 & 90.0   &68,3  & 92.5  &62.6 & 59.9 &42.0&66.6\\
    $\mathcal{H}\mathcal{T}(\theta_{MLM}^{d})$ (TINY)& 72.4 & 78.5  &91.8  & 78.0 & 89.8   &66.0  & 92.5  &62.6 & 59.3 &42.0&63.5\\
    $\mathcal{H}\mathcal{T}(\theta_{MLM}^{u})$ (TINY)& 72.4 & 78.6  &91.8 & 79.0 & 89.8   &65.0  & 91.8  &61.8 & 58.1 &39.2&68.9\\
        HBERT (w) $\theta_{BERT_milmil}$ (TINY)& 70.8 & 77.6  &91.4 & 79.3 & 88.3   &65.8  & 91.9 &58.0 & 56.3 &40.0&59.1 \\ 
           $\mathcal{H}\mathcal{T}(\theta_{MLM}^{u,d})$ (SMALL)& 74.32 & 79.2  & 92.4 & 81.5 & 90.6   &69.4  &92.7 & 64.1 &60.1& 45.0 & 68.2 \\ \hline

      $\mathcal{H}\mathcal{T}(\theta_{GAP}^{d})$ (TINY) & 71.58 & 78.6  & 91.8 & 78.1 & 89.3   & 64.1 & 91.6 & 60.5 & 55.7 & 42.2 & 63.9 \\
       $\mathcal{H}\mathcal{T}(\theta_{GAP}^{u})$ (TINY)&71.52 & 78.5  & 90.9 &79.0  & 88.9 & 66.3  &92.0 & 59.2 &57.5& 39.9 & 63.0 \\\hline

\end{tabular}}

\caption{Performances of all mentioned model with different decoders such as MLP, GRU, CRF  \texttt{SILICONE}. The datasets are grouped by
label type (\texttt{DA} vs \texttt{E/S}) and order by decreasing size.}
\label{tab:full_results}
\end{center}
\end{table*}

\subsection{Improvement over pre-training}
\begin{figure}
  \centering
  \includegraphics[width=\linewidth]{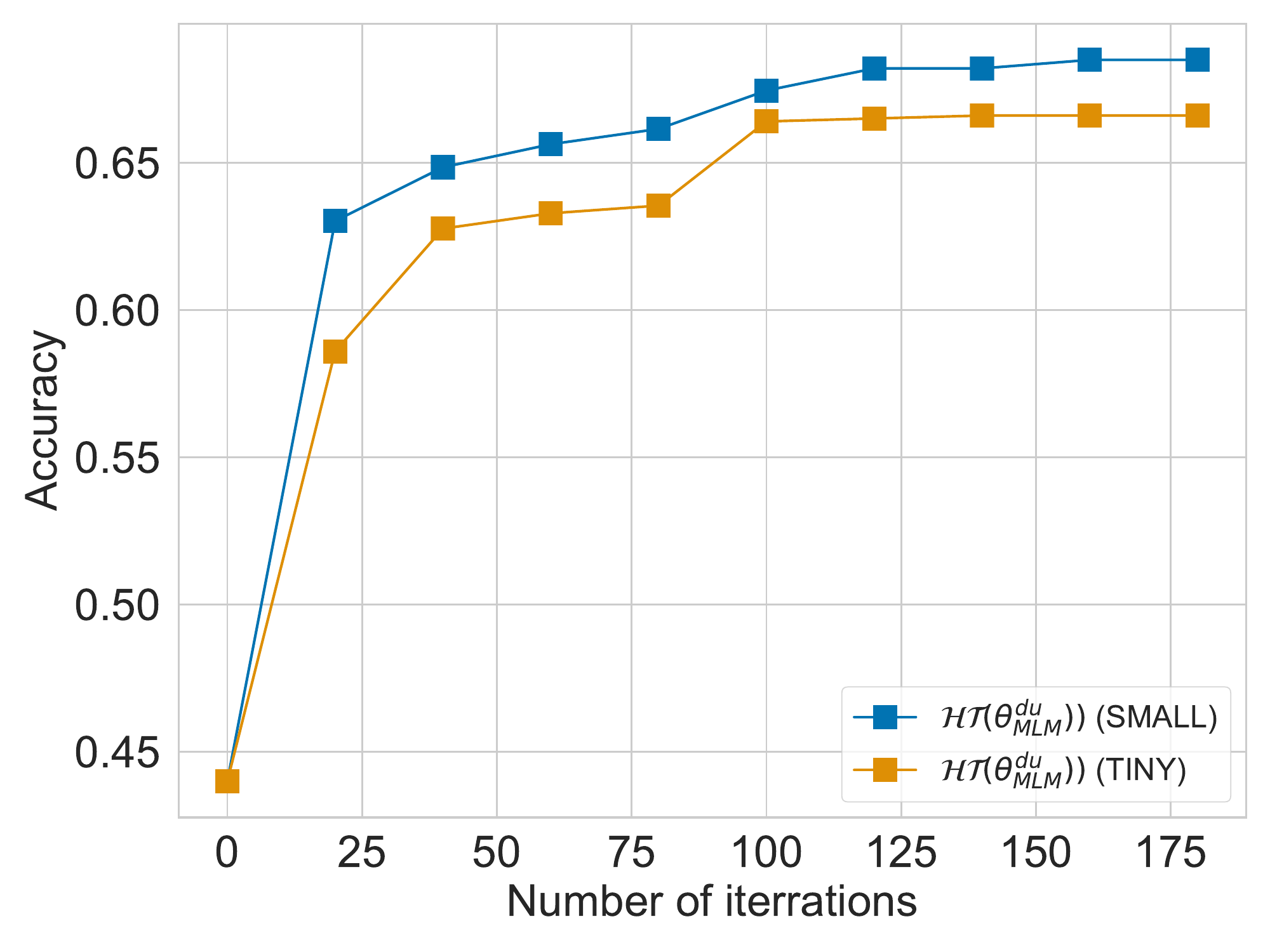}
\caption{Illustration of improvement of accuracy during pre-training stage on \texttt{SEM} for both a \texttt{TINY} and \texttt{SMALL} model.}
  \label{fig:SEMAINE}
\end{figure}

In this experiment we illustrate how pre-training improves performance on \texttt{SEM} (see~\autoref{fig:SEMAINE}). As expected accuracy improves when pre-training.  

\subsection{Multi level Supervision for pre-training \texttt{MELD}} 
In this experiment we report results of the experiment mentioned in~\autoref{sssec:multi_sup}. In this experiment we see that the training process seems to be noisier for fractions lower than 40\%. For larger percentages, we observe that including higher supervision (at the dialog level) during pre-training leads to a consistent improvement.
\begin{figure}
  \centering
  \includegraphics[width=0.4\textwidth]{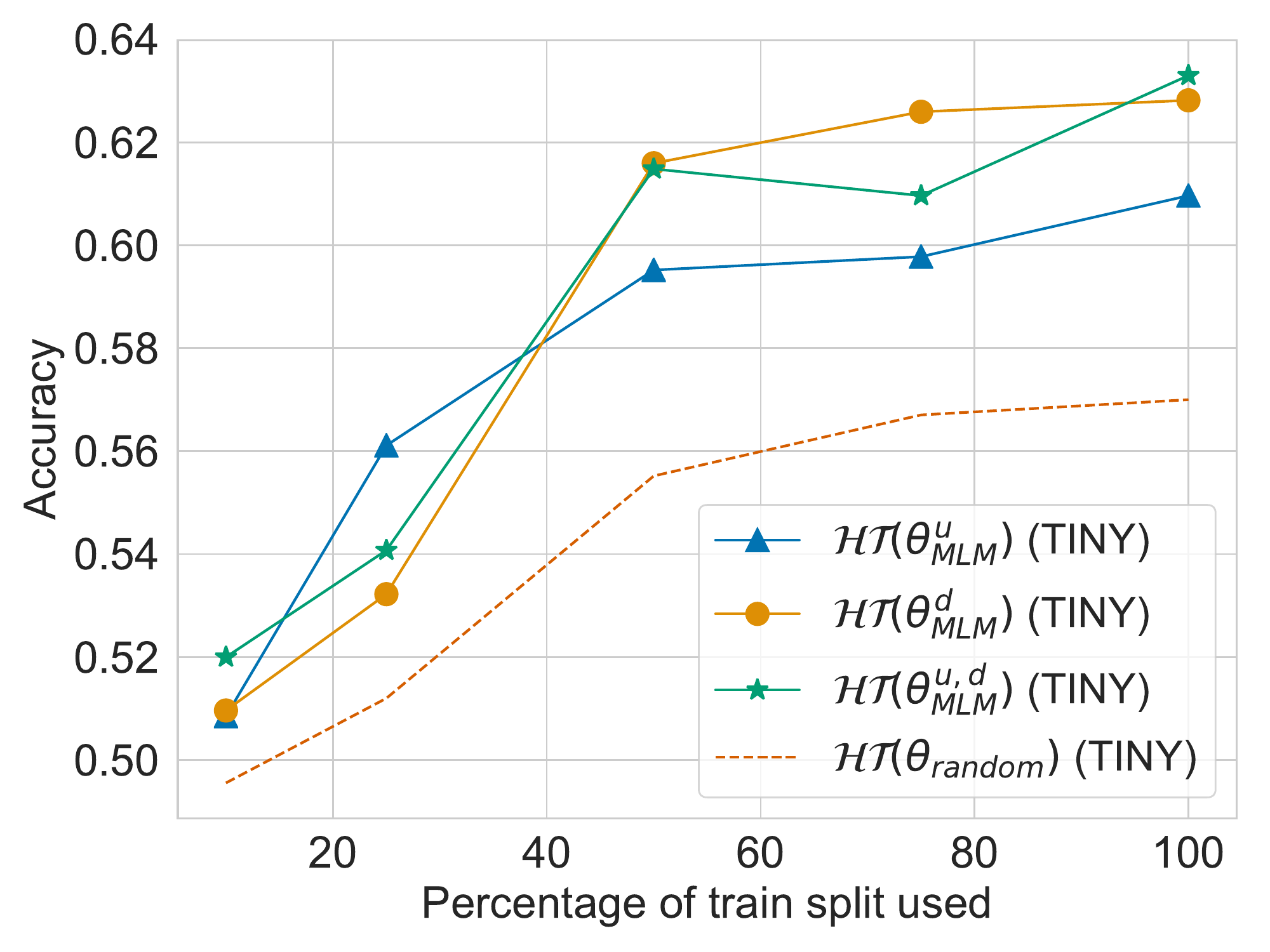}
    \caption{A comparison of different parameters initialisation on $\texttt{MELD}_\texttt{s}$. Training is performed using a different percentage of complete training set. Validation and test set are fixed over all experimentation. Each score  is the averaged accuracy over 10 random runs.}
    \label{fig:exp_split_meld_s}
\end{figure}

\section{Negative Results on \texttt{GAP}}
\label{sec:sup_neg_res}
We briefly describe few ideas we tried to make \texttt{GAP} works at both the utterance and dialog level. We hypothesise that:
\begin{itemize}
    \item giving the same weight to the utterance level and the dialog level (see \autoref{eq:multi_obj}) was responsible of the observed plateau. Different combinations lead to fairly poor improvements.
    \item the limited model capacity was part of the issue. Larger models does not give the expected results.
\end{itemize}

\end{document}